\documentclass[11pt,a4paper]{article}

\usepackage[a4paper,margin=2.35cm]{geometry}
\usepackage{fontspec}
\defaultfontfeatures{Renderer=HarfBuzz}
\usepackage{setspace}
\usepackage{amsmath,amssymb,amsthm,mathtools}
\DeclareMathOperator{\softmaxop}{softmax}
\DeclareMathOperator{\Attnop}{Attn}

\usepackage{graphicx}
\graphicspath{{figures/}}
\usepackage{booktabs,multirow,array,makecell,float,adjustbox,caption,tabularx}
\usepackage{ragged2e}
\newcommand{\best}[1]{\textbf{#1}}

\newenvironment{compacttable}[1][4pt]{\scriptsize\setlength{\tabcolsep}{#1}\renewcommand{\arraystretch}{0.92}}{\normalsize\renewcommand{\arraystretch}{1.0}}
\usepackage[section]{placeins}
\usepackage{enumitem}
\usepackage[dvipsnames]{xcolor}
\definecolor{deepgreen}{HTML}{0B5D43}
\definecolor{lightabstract}{HTML}{F3F5F4}
\definecolor{softgreen}{HTML}{DCEBE5}
\definecolor{linkgreen}{HTML}{0A6B4A}
\usepackage{titlesec}
\titleformat{\section}{\color{deepgreen}\Large\sffamily\bfseries}{\thesection}{0.8em}{}
\titleformat{\subsection}{\color{deepgreen}\large\sffamily\bfseries}{\thesubsection}{0.8em}{}
\titleformat{\subsubsection}{\color{deepgreen}\normalsize\sffamily\bfseries}{\thesubsubsection}{0.8em}{}
\usepackage{tcolorbox}
\tcbuselibrary{breakable,skins}
\usepackage[numbers,sort&compress]{natbib}
\usepackage[colorlinks=true,linkcolor=linkgreen,citecolor=linkgreen,urlcolor=linkgreen]{hyperref}
\usepackage{fancyhdr}
\usepackage{microtype}
\usepackage[dvipsnames]{xcolor}
\renewcommand{\headrulewidth}{0.5pt}
\renewcommand{\headrule}{\hbox to\headwidth{\color{deepgreen}\leaders\hrule height \headrulewidth\hfill}}

\begin{document}

\begin{titlepage}
\thispagestyle{empty}
\noindent\includegraphics[width=0.1\textwidth]{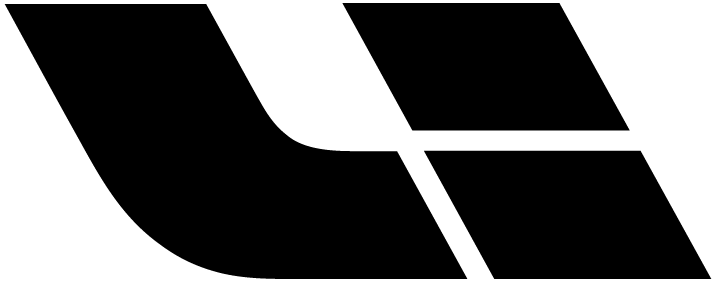}\par
\vspace{0.05em}
{\color{deepgreen}\rule{\textwidth}{0.8pt}}

\begin{center}
{\sffamily\bfseries\fontsize{21}{26}\selectfont
\textcolor{deepgreen}{VLAFlow}: A Unified Training Framework for Vision-Language-Action Models via Co-training and Future Latent Alignment\par}

\vspace{1.0em}

{\sffamily\small
Guoyang Xia\textsuperscript{1,2,*}\quad
Fengfa Li\textsuperscript{1,*}\quad
Hongjin Ji\textsuperscript{1,3}\quad
Lei Ren\textsuperscript{1,\ensuremath{\dagger},\ensuremath{\ddagger}}\quad
Fangxiang Feng\textsuperscript{2,\ensuremath{\ddagger}}\quad
Kun Zhan\textsuperscript{1}\\
Yan Xie\textsuperscript{1}
\par}

\vspace{0.55em}

{\sffamily\footnotesize

\textsuperscript{1}Li Auto Inc.\quad
\textsuperscript{2}School of Artificial Intelligence, Beijing University of Posts and Telecommunications\\
\textsuperscript{3}The Chinese University of Hong Kong, Shenzhen
\par}

\vspace{0.35em}

{\sffamily\footnotesize
\textsuperscript{*}Equal contribution.\quad
\textsuperscript{\ensuremath{\dagger}}Project leader.\quad
\textsuperscript{\ensuremath{\ddagger}}Corresponding author.
\par}

\end{center}

\vspace{0.8em}

\begin{tcolorbox}[
  enhanced,
  breakable,
  colback=lightabstract,
  colframe=deepgreen,
  boxrule=0.8pt,
  arc=4pt,
  left=13pt,right=13pt,top=12pt,bottom=12pt,
  width=\textwidth
]
\begingroup\small\setstretch{1.03}
\begin{center}
{\sffamily\bfseries\large\textcolor{deepgreen}{Abstract}}
\end{center}
\vspace{0.1em}

Vision-language-action models (VLAs) have recently advanced robotic manipulation, yet the effects of different robot-data pre-training paradigms remain difficult to compare because existing models often differ in architecture, data, action space, and evaluation protocol. We present \textbf{VLAFlow} (\textbf{Vision-Language-Action Flow}), a unified flow-matching framework for controlled comparison of VLA training objectives. Using a heterogeneous robot corpus, \textbf{OXEMix}, containing approximately $5{,}000$ hours of data from DROID, OpenX-Embodiment, OpenX-Augmented, and RoboCOIN, we evaluate four paradigms under the same $\pi_0$-style architecture, shared VLM backbone, action expert, and $14$-dimensional action space: action-only modeling (\textbf{MindPI}), language-supervised co-training (\textbf{MindLPI}), future latent alignment (\textbf{MindWPI}), and their combination (\textbf{MindLWPI}). Experiments on LIBERO, LIBERO-Plus, and SimplerEnv show that action-only pre-training is sensitive to heterogeneous data. In contrast, language supervision helps preserve vision-language generalization, while future latent alignment improves state-transition and action-outcome modeling. By combining both signals, MindLWPI achieves the most stable overall transfer performance across benchmarks. These results suggest a \emph{meta-action space} view: language and future latent representations provide complementary intermediate constraints that make heterogeneous action supervision smoother and more transferable.

\vspace{0.5em}
\par
\vspace{0.5em}
\noindent\textbf{Code:} 
\href{https://github.com/MindVLA-Team/VLAFlow}{\textcolor{deepgreen}{https://github.com/MindVLA-Team/VLAFlow}}\par
\vspace{0.5em}
\noindent\textbf{Project Page:} 
\href{https://mindvla-team.github.io/VLAFlow}{\textcolor{deepgreen}{https://mindvla-team.github.io/VLAFlow}}\par

\endgroup
\end{tcolorbox}
\vfill
\end{titlepage}

\setcounter{page}{1}

\section{Introduction}

Vision-language-action models (VLAs) have recently made rapid progress in robot learning and embodied intelligence. With the development of large-scale robot data, vision-language models, and continuous action generation methods, increasingly many pre-trained VLA foundation models have demonstrated strong cross-task generalization. For example, models such as LingBot-VLA and ABot-M0 leverage large quantities of high-quality robot data for pre-training and achieve significant performance gains on multiple manipulation tasks, providing partial evidence for the scaling potential of VLA models \citep{wu2026lingbot,yang2026abot}. However, as both data and model scale continue to grow, the cost of pre-training also rises rapidly. How to design more effective training paradigms that can fully exploit heterogeneous robot data and transfer stably to downstream tasks remains insufficiently understood.

Existing studies suggest that VLA pre-training does not yield stable gains simply by increasing data scale~\citep{ye2026starvlaalphareducingcomplexityvisionlanguageaction}. The conventional action-modeling paradigm represented by $\pi_0$ usually predicts action chunks directly from visual observations and language instructions, and uses the same objective during downstream fine-tuning \citep{black2024pi0}. Under such action-only modeling, however, the pre-training data distribution often plays a decisive role in downstream performance. When the pre-training data and downstream task differ substantially in robot embodiment, action space, sampling frequency, or task semantics, the model may suffer negative transfer. This phenomenon indicates that the key to VLA pre-training lies not only in data scale, but also in whether the pre-training objective can learn transferable intermediate representations from highly heterogeneous data \citep{chen2018gradnorm,hejna2025remix}.

Based on this observation, we focus on a central question: under similar data distributions and the same model architecture, \textbf{how do different VLA training paradigms affect downstream transfer performance?} To answer this question, we categorize representative VLA training methods into four groups. The first is the action-only modeling paradigm, represented by $\pi_0$, where both pre-training and downstream fine-tuning use action-chunk prediction as the main objective \citep{black2024pi0}. The second is the VLM co-training paradigm, such as $\pi_{0.5}$, LAP, and JoyAI-RA~0.1, where the VLM is additionally asked to generate subtask goals, action descriptions, or discretized action expressions, thereby introducing high-level action-intent supervision through language space \citep{black2025pi05,li2026lap,joyai2026ra}. The third is the future latent feature alignment paradigm, such as Being-H0.7 and LDA-1B, where the model predicts or aligns the latent representation of future frames while predicting actions, enabling feature-level future imagination \citep{beingh2026,jiang2026lda}. The fourth is the combined paradigm explored in this report, where language action descriptions and future latent supervision are introduced simultaneously so that language space and visual latent space jointly constrain action representation learning.

To fill the gap in controlled comparison, we propose \textbf{VLAFlow}, or \textbf{Vision-Language-Action Flow}. VLAFlow does not refer to a single model variant; rather, it is a unified evaluation framework for VLA training paradigms. ``Flow'' refers both to the flow-matching action-modeling mechanism in the shared architecture and to how different supervision signals---low-level actions, language intent, and future latent states---flow into the same VLA action-generation framework and ultimately affect downstream robotic control. By placing different pre-training objectives under the same architecture, action space, and evaluation protocol, VLAFlow enables a more direct analysis of how the training paradigm itself affects transfer performance.

To reduce the experimental cost of large-scale pre-training while maintaining representative data diversity, we construct OXEMix, a medium-scale VLA pre-training corpus of approximately $5{,}000$ hours. The corpus is built from widely used open-source robot datasets, including DROID, OpenX-Embodiment, OpenX-Augmented, and RoboCOIN \citep{khazatsky2024droid,openx2023,ji2025oxe,wu2025robocoin}. In model design, we adopt a $\pi_0$-style architecture consistent with recent mainstream approaches and model continuous actions using flow-matching loss, thereby controlling as many confounding factors from architecture and optimization as possible \citep{lipman2022flow,black2024pi0}. On this basis, we design and compare four training paradigms: \textbf{MindPI}, \textbf{MindLPI}, \textbf{MindWPI}, and \textbf{MindLWPI}.

Specifically, MindPI corresponds to the action-only modeling paradigm and primarily optimizes a continuous-action flow-matching loss during pre-training. MindLPI corresponds to the VLM co-training paradigm and introduces LAP-style~\citep{li2026lap} action description templates during pre-training, allowing the model to learn both language-level action-intent expression and action modeling; during downstream fine-tuning, we retain only the action-modeling objective to avoid lowering the control frequency with long language generation. MindWPI corresponds to the future latent feature alignment paradigm: it uses V-JEPA~2 as a frozen latent feature extractor, feeds current-frame latent features as additional context to the action expert, and requires the model to predict the latent representation of future frames while still using flow matching for action modeling \citep{bardes2025vjepa2}. MindLWPI further combines MindLPI and MindWPI by using action loss, language action-description loss, and future latent loss during pre-training, and applies average pooling to compress visual latent tokens, reducing token overhead during joint training and inference.

We conduct systematic evaluations on the downstream benchmarks LIBERO, LIBERO-Plus, and SimplerEnv, together with extensive ablations to analyze the mechanisms behind different training paradigms \citep{liu2023libero,li2025liberoplus,li2024simpler}. LIBERO is already close to saturation and thus better serves as a sanity check for basic capability. LIBERO-Plus zero-shot perturbation evaluation reveals the trade-off between the degree of VLM intervention and vision-language generalization. SimplerEnv further amplifies cross-embodiment transfer differences caused by different pre-training objectives. Overall, action-only full-parameter pre-training tends to produce negative transfer on heterogeneous data; language supervision and future latent supervision mitigate this issue from the perspectives of high-level intent and state transition, respectively; and MindLWPI integrates the two to achieve the most stable overall performance across LIBERO, LIBERO-Plus, and SimplerEnv.

We argue that MindLPI, MindWPI, and MindLWPI outperform action-only modeling because they provide intermediate representations for action learning. In highly heterogeneous pre-training data, robot platforms, action definitions, sampling frequencies, and task semantics differ substantially. Relying only on low-dimensional action supervision makes it difficult to form a stable ``meta-action'' representation. In contrast, language descriptions explicitly express high-level action intent, while future latent features capture action-induced state changes. When combined, these two forms of supervision better smooth the optimization space between action intent and real control signals, helping the model form more generalizable action representations.

The main contributions of this report are summarized as follows:
\begin{enumerate}[leftmargin=2em,itemsep=0.35em]
    \item \textbf{We propose VLAFlow, a unified flow-matching framework for controlled comparison of VLA training paradigms.} Using approximately $5{,}000$ hours of medium-scale heterogeneous robot data, we conduct a fair comparison of action-only modeling, language-supervised co-training, future latent alignment, and their combination under a unified model architecture, action space, and evaluation protocol.
    \item \textbf{We show that action-only modeling is sensitive to pre-training settings on heterogeneous robot data and may incur negative transfer.} Experiments indicate that full-parameter action-only pre-training can damage transfer ability on downstream tasks with large distribution gaps. Freezing the VLM can partially preserve vision-language generalization, but does not fully exploit robot data to learn action-related state changes.
    \item \textbf{We validate the complementarity between language supervision and future latent alignment.} MindLPI provides high-level action-intent supervision through language action descriptions, MindWPI introduces state-transition constraints through future visual latent prediction, and MindLWPI combines the two to obtain more stable transfer performance across multiple benchmarks.
    \item \textbf{We introduce a meta-action-space interpretation.} Language space and future visual latent space provide intermediate constraints for heterogeneous action supervision from the perspectives of high-level intent and state transition, helping explain transfer differences among VLA training paradigms on heterogeneous robot data.
\end{enumerate}

\section{Related Work}

\subsection{Vision-Language-Action Models and Training Paradigms}

Vision-language-action (VLA) models aim to unify visual perception, language understanding, and robot control within an end-to-end framework. Early works such as RT-1 and RT-2 demonstrated the effectiveness of large-scale Transformer-based robot policy learning and the potential for transferring vision-language pre-training knowledge to robotic control \citep{brohan2022rt1,zitkovich2023rt2}. Later, open-source models such as OpenVLA and Octo further advanced VLA architectures and training paradigms: OpenVLA builds a general VLA baseline on Open X-Embodiment using a large language model and visual encoders, whereas Octo emphasizes a lightweight Transformer and a diffusion policy head for efficient fine-tuning and edge deployment \citep{kim2024openvla,ghosh2024octo,openx2023}. More recently, $\pi_0$ introduced flow matching into continuous action generation and combined it with a VLM backbone to enable high-frequency and fine-grained robot control, becoming an important representative VLA architecture \citep{black2024pi0}. $\pi_{0.5}$ represents a VLM co-training paradigm that goes beyond action imitation by incorporating heterogeneous supervision sources, including high-level semantic subtask prediction, verbal instructions, cross-embodiment robot data, and web multimodal data. Its two-stage recipe first uses FAST-tokenized discrete actions for scalable pre-training, and then adds a flow-matching action expert in post-training to recover fine-grained continuous control \citep{black2025pi05}.

Existing VLA training paradigms can be broadly divided into three categories, and can be further extended to combined paradigms. The first is the action-only modeling paradigm represented by $\pi_0$, whose core objective is to predict action chunks conditioned on visual observations and task instructions \citep{black2024pi0}. This paradigm has a simple objective and can be scaled to large heterogeneous datasets, but it is also highly dependent on the pre-training data distribution. When the pre-training data differ substantially from the downstream task, action-only modeling can suffer negative transfer. The second is the VLM co-training paradigm. These methods introduce intermediate supervision in language space in addition to action modeling, such as subtask goals, action descriptions, or discretized action tokens. Works such as $\pi_{0.5}$, LAP, and JoyAI-RA~0.1 all exploit language representations to improve modeling of task semantics and action intent to different degrees \citep{black2025pi05,li2026lap,joyai2026ra}. The third is the future latent feature alignment paradigm. These methods require the model to predict or align representations of future observations in latent space while predicting actions, giving VLA models feature-level future imagination capability \citep{beingh2026,jiang2026lda,sun2026vlajepa,miao2026jepavla}. The MindLWPI paradigm in this report can be viewed as a combination of language-supervised co-training and future latent alignment, designed to study whether the two intermediate supervision signals are complementary.

The development of these training paradigms relies heavily on large-scale robot datasets. Open X-Embodiment aggregates large-scale demonstration trajectories across many robot platforms and is currently one of the most widely used cross-embodiment pre-training corpora \citep{openx2023}. DROID collects diverse manipulation data in real homes and offices \citep{khazatsky2024droid}. RoboCOIN and RoboMIND further emphasize cross-platform, cross-task, and multi-embodiment data integration \citep{wu2025robocoin,wu2024robomind}. These datasets provide rich data foundations for VLA pre-training, but their high heterogeneity in sampling frequency, action space, task semantics, and robot embodiment also makes the design of effective pre-training objectives a key challenge.

\subsection{World Models and Future Prediction in Robot Learning}

World models aim to learn environment dynamics so that an agent can internally predict the future outcomes of actions and use such predictions to support decision making. In robot learning, early methods often adopt a decoupled ``predict-then-act'' framework. For example, UniPi and VidMan predict future observations through video generation models and then recover actions using inverse dynamics models \citep{du2023unipi,wen2024vidman}. Later works such as Cosmos Policy and DreamZero attempt to unify future visual prediction and action modeling within a shared generative framework \citep{kim2026cosmos,wang2026dreamzero}. However, pixel-level prediction typically incurs high computational cost and is susceptible to redundant visual details and long-horizon error accumulation.

Joint-Embedding Predictive Architecture (JEPA) provides a more efficient alternative: instead of reconstructing pixels directly, it predicts features of future or masked regions in representation space \citep{assran2023self}. V-JEPA extends this idea to video representation learning, while V-JEPA~2 further demonstrates the potential of scaled self-supervised video pre-training and lightweight robot-data post-training for robot control \citep{bardes2024vjepa,bardes2025vjepa2}. Recent VLA-JEPA and JEPA-VLA works introduce the predictive representations of V-JEPA~2 into VLA models, suggesting that future visual latent representations can provide useful dynamics priors for policy learning \citep{sun2026vlajepa,miao2026jepavla}.

Compared with explicit video prediction, latent-space world modeling is better suited for VLA pre-training. Being-H0.7 uses alignment between a prior branch and a posterior branch in latent space, allowing the model to obtain future-aware implicit reasoning at inference time using only current observations \citep{beingh2026}. LDA-1B unifies visual forecasting and action generation in a DINO latent space to improve the use of heterogeneous embodied data \citep{jiang2026lda}. Together, these works indicate that future latent prediction can improve the model's understanding of environment dynamics and action outcomes while maintaining inference efficiency. Building on this line of work, this report systematically compares different VLA training paradigms and validates the contribution of future latent alignment, as well as its combination with language supervision, under controlled data and architecture settings.

\section{Method}

To fairly evaluate the effect of training objectives on downstream VLA transfer, we construct a unified VLAFlow framework and instantiate four training paradigms within it: MindPI, MindLPI, MindWPI, and MindLWPI. Rather than proposing a single new architecture, the central goal of this report is to control architecture, data, action space, and optimization settings so that the main differences among methods are concentrated in the training supervision signals. This section first summarizes the controlled comparison protocol, and then introduces the shared VLAFlow architecture and the four training paradigms. Additional implementation details, including key-value cache sharing, attention masks, action verbalization rules, latent compression, training hyperparameters, and LoRA configurations, are provided in Appendices~\ref{app:implementation}--\ref{app:lora}.

\subsection{Controlled Comparison Protocol}
\label{subsec:controlled_protocol}

This report asks the following question: under approximately the same data distribution and model architecture, do different VLA training objectives lead to different downstream transfer behaviors? To answer this, we adopt four control principles.

\textbf{Same model backbone.} All four paradigms use the same vision-language model (VLM) as the multimodal context encoder, followed by a $\pi_0$-style continuous action expert. The action expert models future action chunks with flow matching and does not use an additional discrete action decoder as the downstream control path.

\textbf{Same action space.} All pre-training data are mapped into a $14$-dimensional action space. Each arm contributes a $7$-dimensional action composed of end-effector translation increments, rotation increments, and gripper state; two arms are concatenated into $14$ dimensions. Single-arm data are incorporated into the unified space by zero-padding the unused arm and applying an action-validity mask.

\textbf{Same data sources and training budget.} The four paradigms share OXEMix, a medium-scale open-source robot-data mixture of approximately $5{,}000$ hours, whose main sources include DROID, OpenX-Embodiment, OpenX-Augmented, and RoboCOIN. Except for data fields required by specific auxiliary supervision, all paradigms use the same data sampling strategy, number of training steps, optimizer, and learning-rate configuration.

\textbf{Same downstream evaluation protocol.} All pre-trained checkpoints are adapted and evaluated under the same benchmark-specific protocols. Models are fine-tuned on standard LIBERO and then evaluated on both LIBERO and LIBERO-Plus, with LIBERO-Plus serving exclusively as a zero-shot robustness benchmark without any additional adaptation. SimplerEnv models are fine-tuned on the corresponding downstream robot data. We also train downstream models initialized without robot-data pre-training as baselines, in order to measure positive or negative transfer induced by different training paradigms.

Table~\ref{tab:paradigm_compare} summarizes the main differences among the four paradigms. They share the same inputs, action expert, and downstream control form; the only differences are whether language supervision, future latent supervision, or both are introduced during training.

\begin{table}[t]
\centering
\small
\setlength{\tabcolsep}{5pt}
\renewcommand{\arraystretch}{1.2}
\caption{Controlled comparison of four VLA training paradigms.}
\label{tab:paradigm_compare}

\begin{tabularx}{\linewidth}{@{}l >{\RaggedRight\arraybackslash}X c c >{\RaggedRight\arraybackslash}X@{}}
\toprule
\textcolor{ForestGreen}{\textbf{Paradigm}} &
\textcolor{ForestGreen}{\textbf{Auxiliary supervision}} &
\textcolor{ForestGreen}{\textbf{PT loss}} &
\textcolor{ForestGreen}{\textbf{FT loss}} &
\textcolor{ForestGreen}{\textbf{Main role}} \\
\midrule
MindPI &
- &
$\mathcal{L}_{\mathrm{act}}$ &
$\mathcal{L}_{\mathrm{act}}$ &
Action-only transfer baseline. \\

MindLPI &
language &
$\mathcal{L}_{\mathrm{act}},\mathcal{L}_{\mathrm{lang}}$ &
$\mathcal{L}_{\mathrm{act}}$ &
Injects high-level action intent through language supervision. \\

MindWPI &
future latent &
$\mathcal{L}_{\mathrm{act}},\mathcal{L}_{\mathrm{lat}}$ &
$\mathcal{L}_{\mathrm{act}},\mathcal{L}_{\mathrm{lat}}$ &
Regularizes learning with future-state prediction. \\

MindLWPI &
language + future latent &
$\mathcal{L}_{\mathrm{act}},\mathcal{L}_{\mathrm{lat}},\mathcal{L}_{\mathrm{lang}}$ &
$\mathcal{L}_{\mathrm{act}},\mathcal{L}_{\mathrm{lat}}$ &
Combines intent and state-transition constraints. \\
\bottomrule
\end{tabularx}
\end{table}

\begin{figure}[t]
    \centering
    \includegraphics[width=0.8\linewidth]{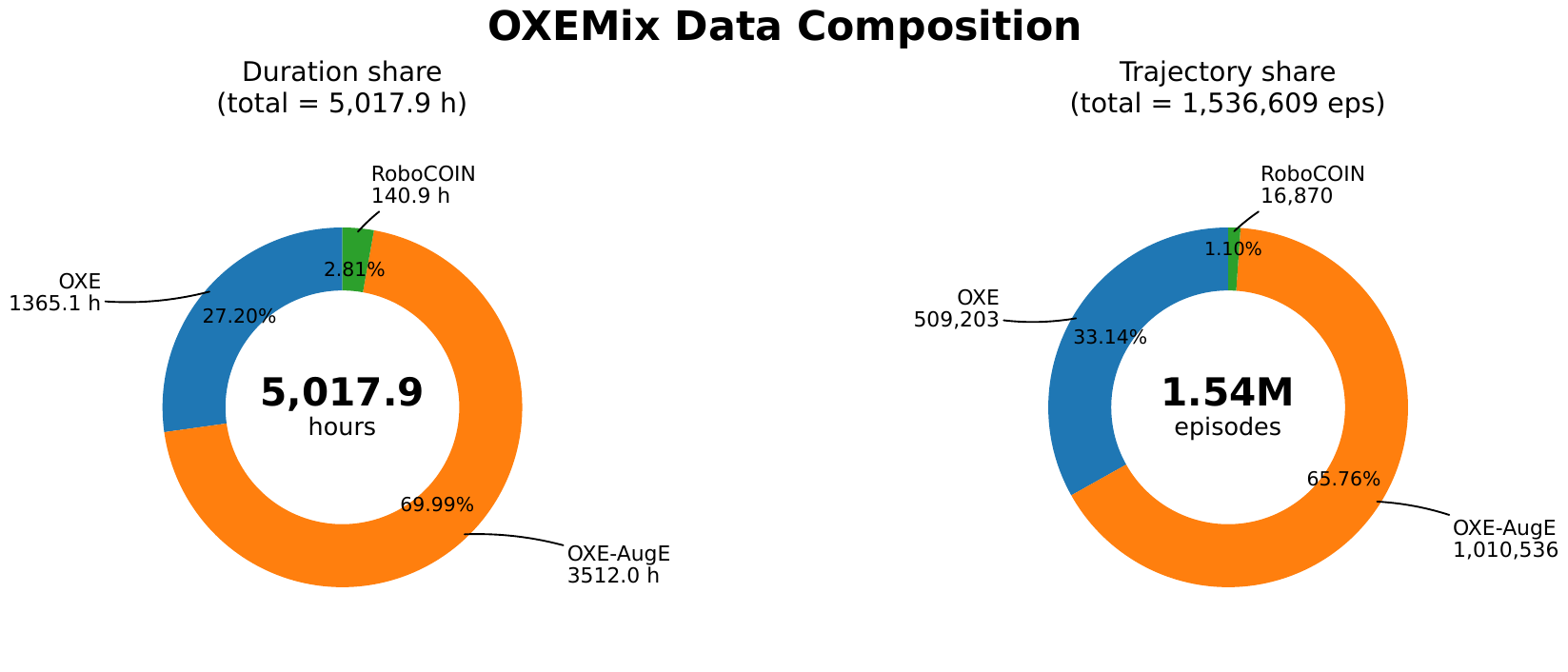}
    \caption{Composition of the OXEMix pre-training corpus in terms of duration and trajectory counts.}
    \label{fig:oxe_mix_composition}
\end{figure}

\subsection{Shared VLAFlow Architecture}
\label{subsec:shared_architecture}

\begin{figure}[!t]
  \centering
  \includegraphics[width=\linewidth]{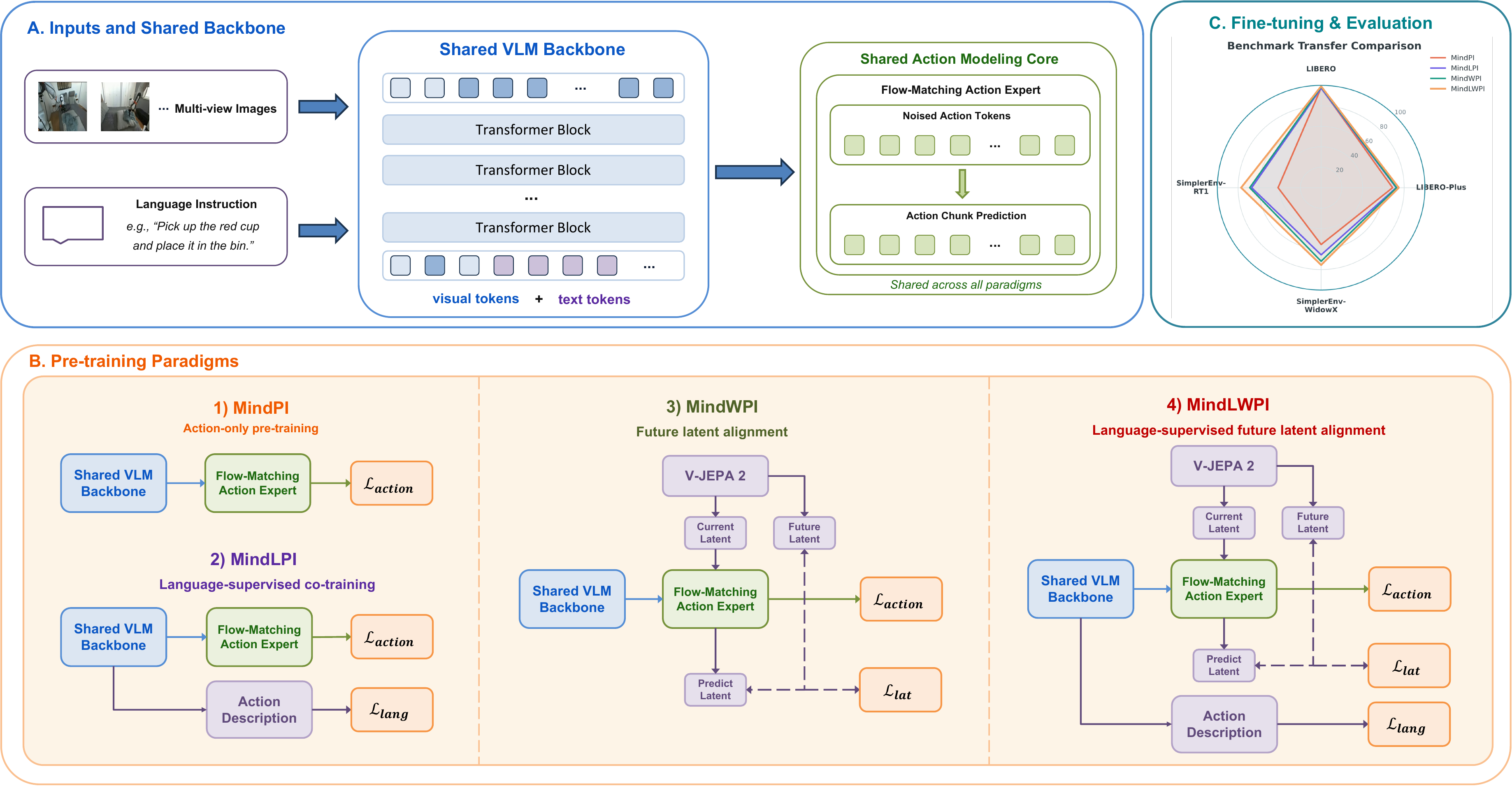}
  \caption{\textbf{Overview of VLAFlow.} VLAFlow conducts controlled comparison of different VLA training paradigms under a unified architecture, data mixture, and evaluation setting. \textbf{A. Inputs and shared backbone.} The model takes multi-view visual observations and a language instruction as input, encodes visual and text tokens with a shared VLM backbone, and passes multimodal context to a flow-matching action expert through KV-cache sharing for continuous action-chunk prediction. \textbf{B. Pre-training paradigms.} The main diagram shows three basic paradigms (MindPI, MindLPI, MindWPI) and their combination MindLWPI: MindPI uses only action-modeling loss; MindLPI introduces language action-description supervision in addition to the action loss; MindWPI introduces future latent alignment by using V-JEPA~2 to extract current-frame and future-frame latent features and requiring the model to predict future latent representations. MindLWPI, the combined extension proposed in this report, integrates MindLPI and MindWPI; its structure and losses are described in the main text. \textbf{C. Fine-tuning and evaluation.} All pre-trained models follow the same benchmark-specific downstream protocols. Models fine-tuned on standard LIBERO are evaluated on LIBERO and, without further adaptation, on LIBERO-Plus for zero-shot robustness; SimplerEnv models are fine-tuned on the corresponding downstream robot data.}
  \label{fig:vlaflow_main}
\end{figure}

VLAFlow uses a two-part structure: a VLM backbone plus an action expert. Given multi-view image observations and a language instruction, the VLM first encodes multimodal context; the action expert then generates a continuous action chunk of length $T$ conditioned on this context. We set $T=16$ by default. To avoid confounding from architectural differences, MindPI, MindLPI, MindWPI, and MindLWPI all use the same shared architecture.

\textbf{VLM context encoding.} The VLM receives images and the language instruction and outputs multi-layer multimodal representations. In our implementation, we use Qwen3-VL-4B-Instruct as the VLM backbone \citep{bai2025qwen3vl}. The action expert does not re-encode images; instead, it reuses the multi-layer context representations from the VLM as conditioning for action generation.

\textbf{Continuous action expert.} The action expert is a diffusion Transformer (DiT) decoder that predicts the flow-matching velocity field for future action chunks \citep{peebles2023dit}. Unlike designs that independently read VLM representations through cross-attention, our implementation uses layer-wise key-value cache sharing, allowing the action expert to reuse multimodal context from different depths of the VLM. This design is shared by all four paradigms and is not a comparison variable; the full formulation is given in Appendix~\ref{app:kv_cache}.

\textbf{Flow-matching action modeling.} Let the ground-truth action chunk be $\mathbf{a}$, the noise be $\boldsymbol{\epsilon}\sim\mathcal{N}(\mathbf{0},\mathbf{I})$, and the continuous time be $t\in[0,1]$. The noised action is defined as
\begin{equation}
\mathbf{x}_{t}=(1-t)\boldsymbol{\epsilon}+t\mathbf{a}.
\end{equation}
The action expert predicts a velocity field $\mathbf{v}_{\theta}(\mathbf{x}_{t},t,\mathbf{c})$ conditioned on the VLM context $\mathbf{c}$, with $\mathbf{a}-\boldsymbol{\epsilon}$ as the target:
\begin{equation}
\mathcal{L}_{\mathrm{act}}
=\mathbb{E}_{t,\boldsymbol{\epsilon}}
\left[\left\|\mathbf{v}_{\theta}(\mathbf{x}_{t},t,\mathbf{c})-(\mathbf{a}-\boldsymbol{\epsilon})\right\|^{2}\right].
\label{eq:act_loss_main}
\end{equation}
When an action-validity mask is available, the loss is computed only over valid action dimensions. At inference time, the model starts from Gaussian noise and generates action chunks through a small number of Euler integration steps. All paradigms deploy only continuous action denoising and do not perform autoregressive language generation, so auxiliary training tasks do not affect control frequency. This formulation follows the general flow-matching framework and recent VLA flow policies \citep{lipman2022flow,black2024pi0}.

\subsection{MindPI: Action-Only Pre-training}
\label{subsec:mindpi_main}

MindPI corresponds to the action-only modeling paradigm represented by $\pi_0$. Its training input consists of the current visual observation, language instruction, and noised action chunk; its only supervision signal is the future action chunk. Both pre-training and downstream fine-tuning optimize the flow-matching action loss in Eq.~\eqref{eq:act_loss_main}:
\begin{equation}
\mathcal{L}_{\mathrm{MindPI}}=\mathcal{L}_{\mathrm{act}}.
\end{equation}
MindPI is the key baseline of this report. Because it introduces no language description, future state, or other intermediate supervision, the model must rely entirely on low-dimensional action labels to learn transferable representations from heterogeneous robot data. Therefore, MindPI can be used to test the training stability of action-only modeling across robot embodiments, sampling frequencies, and action spaces.

\subsection{MindLPI: Language-Supervised VLM Co-training}
\label{subsec:mindlpi_main}

MindLPI corresponds to the VLM co-training paradigm. Its core idea is to introduce intermediate supervision in language space in addition to action modeling, so that the VLM learns to encode high-level action intent in linguistic representations. Specifically, we convert action chunks into LAP-style action-description text and use it as an autoregressive training target for the VLM \citep{li2026lap}. The language supervision in the current experiments is solely derived from action-description templates.

The MindLPI pre-training objective consists of the continuous action loss $\mathcal{L}_{\mathrm{act}}$ and the language action-description loss $\mathcal{L}_{\mathrm{lang}}$:
\begin{equation}
\mathcal{L}_{\mathrm{MindLPI}}
=\mathcal{L}_{\mathrm{act}}+\lambda_{\mathrm{lang}}\mathcal{L}_{\mathrm{lang}}.
\end{equation}
Here $\mathcal{L}_{\mathrm{lang}}$ is the standard autoregressive cross-entropy loss, and we set $\lambda_{\mathrm{lang}}=0.1$ in the current experiments. Language action descriptions can take the form of discretized integer action sequences or natural-language action templates; details are provided in Appendix~\ref{app:language_action}. In ablations, we compare whether the action loss is backpropagated into the VLM, and the results show that removing gradient truncation is more suitable as the main setting.

During downstream fine-tuning, MindLPI keeps only the continuous action loss $\mathcal{L}_{\mathrm{act}}$ and no longer generates language action descriptions. This design matches deployment requirements: language supervision shapes representations only during pre-training, while closed-loop control does not introduce additional autoregressive language latency.

\subsection{MindWPI: Future Latent Feature Alignment}
\label{subsec:mindwpi_main}

MindWPI corresponds to the future latent feature alignment paradigm. Its core hypothesis is that effective VLA pre-training should learn not only a mapping from current observations to actions, but also state changes that actions may induce. Compared with pixel-level future-frame reconstruction, latent-space prediction avoids modeling visual details weakly related to control, such as texture, lighting, and background, while retaining compact information about dynamics, contact, and task progress.

Specifically, we use a frozen V-JEPA~2 model as the latent feature extractor \citep{bardes2025vjepa2}. Given current and future frames, we extract latent representations $\mathbf{z}_{\mathrm{cur}}$ and $\mathbf{z}_{\mathrm{fut}}$. The current latent representation is encoded and used as additional tokens for the action expert; while predicting the action velocity field, the action expert also predicts the future latent representation $\hat{\mathbf{z}}_{\mathrm{fut}}$ through a latent decoder. The future latent loss is defined as
\begin{equation}
\mathcal{L}_{\mathrm{lat}}
=\left\|\hat{\mathbf{z}}_{\mathrm{fut}}-\mathbf{z}_{\mathrm{fut}}\right\|^{2}.
\end{equation}
The joint objective of MindWPI is
\begin{equation}
\mathcal{L}_{\mathrm{MindWPI}}
=\mathcal{L}_{\mathrm{act}}
+\lambda_{\mathrm{lat}}\mathcal{L}_{\mathrm{lat}}.
\end{equation}

To prevent future latent prediction from taking a shortcut through action tokens, we use a structured attention mask: latent tokens can predict the future only from the current observation and language context, and cannot attend to noised action tokens; action tokens, however, can attend to latent tokens and use them as predictive visual context. The full attention-mask form is given in Appendix~\ref{app:mask}.

During downstream fine-tuning, MindWPI continues to retain both action loss and future latent loss. At inference time, the model requires only the current-frame latent representation as condition and outputs both the future latent representation and the action chunk. Therefore, MindWPI uses future prediction as a representation constraint during training without changing the deployment interface for action generation.

\subsection{MindLWPI: Joint Language Supervision and Future Latent Pre-training}
\label{subsec:mindlwpi_main}

MindLWPI (Language-supervised Future Latent Alignment) further combines MindLPI and MindWPI. The motivation is that language action descriptions provide high-level action-intent supervision, while future latent prediction provides action-outcome and state-transition supervision. The two respectively constrain ``what to do'' and ``what the action will change.'' Thus, MindLWPI simultaneously optimizes action modeling, language action description, and future latent prediction during pre-training:
\begin{equation}
\mathcal{L}_{\mathrm{MindLWPI}}
=\mathcal{L}_{\mathrm{act}}
+\lambda_{\mathrm{lat}}\mathcal{L}_{\mathrm{lat}}
+\lambda_{\mathrm{lang}}\mathcal{L}_{\mathrm{lang}}.
\end{equation}
In all reported MindLWPI results, when both the action loss and future latent loss are enabled, their ratio during pre-training is fixed at $1:1$. The language action-description loss weight follows the MindLPI setting, namely $\lambda_{\mathrm{lang}}=0.1$. As in MindLPI, MindLWPI disables the language loss during downstream fine-tuning and retains only the action loss and future latent loss.

Directly using all $256$ V-JEPA~2 latent tokens for both language co-training and future prediction substantially increases the sequence length of the action expert. To reduce inference overhead, MindLWPI uses AvgPool-k4 compression by default: every $4$ adjacent tokens along the latent-token sequence are averaged, reducing $256$ tokens to $64$ tokens. This compression is applied to both the current latent prefix and the future latent target so that the prediction dimensions are consistent. We compare AvgPool and MLP compression, as well as k4 and k16 settings, in ablation experiments. The results show that simple AvgPool-k4 offers a favorable trade-off between performance and computational cost. Therefore, unless otherwise stated, MindLWPI in the main text refers to ``language supervision + future latent alignment + AvgPool-k4 + downstream $\lambda^{\mathrm{ft}}_{\mathrm{lat}}:\lambda^{\mathrm{ft}}_{\mathrm{act}}=0.1:1$.''

\subsection{Training, Fine-tuning, and Efficient Adaptation Protocol}
\label{subsec:training_protocol_main}

The four paradigms share the same pre-training data mixture and optimization budget. Pre-training data are uniformly converted into the LeRobot format and mapped into the $14$-dimensional action space described in Section~\ref{subsec:controlled_protocol} \citep{lerobot2024}. To address sampling imbalance caused by different dataset sizes, we use a sampling strategy that balances dataset scale and trajectory length. Optimizer, learning rate, training steps, and flow-matching sampling details are provided in Appendix~\ref{app:training_details}.

At the downstream stage, we transfer checkpoints from different paradigms to the same evaluation benchmarks and use the same fine-tuning protocol. Whether a pre-trained checkpoint is loaded distinguishes pre-training transfer results from downstream baselines without pre-training. For benchmarks such as LIBERO whose action definitions differ from the pre-training phase, the VLM mainly transfers vision-language representations, while the action expert adapts to the new action space with a higher learning rate.

In addition to full-parameter fine-tuning, we also explore low-rank adaptation (LoRA) to evaluate downstream adaptation under resource constraints \citep{hu2021lora}. LoRA experiments are not a primary variable in the comparison of the four training paradigms, so their implementation details and injection locations are provided in Appendix~\ref{app:lora}; the experimental results are reported in Section~\ref{sec:experiments}.

\section{Experiments}
\label{sec:experiments}

\subsection{Experimental Setup}

\paragraph{Evaluation benchmarks and metrics.}
We evaluate different training paradigms on three downstream benchmarks. \textbf{LIBERO} contains four standard suites: L-Spatial, L-Object, L-Goal, and L-Long. Each suite is evaluated over 500 rollouts, and task success rate (\%) is used as the metric \citep{liu2023libero}. \textbf{LIBERO-Plus} introduces seven types of perturbations on top of standard LIBERO tasks, including camera viewpoint, robot initial state, language instruction, lighting, background texture, sensor noise, and object layout, to evaluate zero-shot robustness \citep{li2025liberoplus}. All models are trained or fine-tuned only on standard LIBERO data and are not adapted on LIBERO-Plus. \textbf{SimplerEnv} contains two robot platforms, WidowX and RT-1 \citep{li2024simpler}. WidowX evaluates four tasks: Stack, Carrot, Spoon, and Eggplant; RT-1 evaluates two visual settings, Visual Matching (VM) and Visual Augmentation (VA). Unless otherwise stated, all results are success rates (\%).

\paragraph{Compared methods.}
To disentangle architectural gains, pre-training gains, and training-objective gains, we compare seven VLAFlow variants: MindPI w/o PT, MindWPI w/o PT, MindPI (Frozen VLM), MindPI (Full PT), MindLPI, MindWPI, and MindLWPI. Here w/o PT means no robot-data pre-training, and Full PT means that both the VLM and action expert are updated during pre-training. MindLWPI by default denotes Language-supervised Future Latent Alignment with AvgPool-k4 latent compression and downstream $\lambda^{\mathrm{ft}}_{\mathrm{lat}}:\lambda^{\mathrm{ft}}_{\mathrm{act}}=0.1:1$.

\paragraph{Checkpoint and downstream-data selection rules.}
For both the standard LIBERO evaluation and the zero-shot LIBERO-Plus evaluation, we uniformly use the same 100k-step checkpoint fine-tuned only on standard LIBERO. For the main SimplerEnv table, we adopt a split fine-tuning strategy: WidowX uses the best checkpoint from Bridge-only fine-tuning, whereas RT-1 uses the best checkpoint from RT-1-only fine-tuning. For many low-cost ablations, we use a mixed SimplerEnv fine-tuning setting with both Bridge and RT-1 data to reduce the cost of full pre-training and per-platform fine-tuning; Section~\ref{subsec:mixed_split} discusses this trade-off explicitly.

\paragraph{Definition of negative transfer.}
We define negative transfer as the case where a model pre-trained on robot data performs worse than its corresponding no-pre-training baseline under the same downstream fine-tuning protocol. This definition is used to analyze whether heterogeneous pre-training data are stably converted into downstream gains under different training paradigms.

\subsection{Controlled Comparison: Overall Transfer Behavior Across Training Paradigms}
\label{subsec:controlled_results}

Table~\ref{tab:controlled_overall_en} summarizes the controlled comparison of major VLAFlow variants on the three downstream benchmarks. This table compares the transfer behavior of different training objectives under the same architecture, action space, and evaluation protocol.

\begin{table}[H]
\centering
\caption{Controlled comparison of VLAFlow training paradigms. For SimplerEnv, WidowX uses Bridge-only fine-tuning results, and RT-1 uses RT-1-only fine-tuning results. MindLWPI uses AvgPool-k4 latent compression and downstream $0.1:1$ FT ratio by default.}
\label{tab:controlled_overall_en}
\scriptsize
\begin{adjustbox}{width=\textwidth}
\begin{tabular}{lccccccc}
\toprule
Method & Robot pre-training & Auxiliary supervision & LIBERO Avg & LIBERO-Plus Total & WidowX Avg & RT-1 VM & RT-1 VA \\
\midrule
\multicolumn{8}{l}{\textit{No robot-data pre-training}} \\
MindPI w/o PT & No & - & 97.0 & 59.9 & 59.6 & 75.7 & 60.4 \\
MindWPI w/o PT & No & future latent & 97.4 & 66.1 & 71.9 & 75.2 & 51.6 \\
\midrule
\multicolumn{8}{l}{\textit{Action-only pre-training}} \\
MindPI (Frozen VLM) & Yes & - & 97.2 & \best{74.9} & 54.4 & 72.7 & 66.0 \\
MindPI (Full PT) & Yes & - & 97.5 & 68.8 & 65.9 & 68.2 & 55.5 \\
\midrule
\multicolumn{8}{l}{\textit{Auxiliary-supervised pre-training}} \\
MindLPI & Yes & language & 97.2 & 72.3 & 65.6 & 74.6 & 59.2 \\
MindWPI & Yes & future latent & 98.5 & 72.6 & 74.5 & \best{86.7} & \best{71.1} \\
MindLWPI & Yes & language + future latent & \best{99.1} & 74.8 & \best{75.5} & 84.4 & 69.8 \\
\bottomrule
\end{tabular}
\end{adjustbox}
\end{table}

LIBERO is already close to saturation, with absolute differences mainly within 1--2 percentage points; it is therefore better suited as an in-distribution capability check. In contrast, zero-shot LIBERO-Plus and SimplerEnv better reveal transfer differences among training paradigms. MindPI (Full PT) improves over the no-pretraining baseline on WidowX but degrades substantially on RT-1 VM/VA, reflecting the instability and potential negative transfer of action-only pre-training under heterogeneous robot data. Among auxiliary-supervised paradigms, MindWPI achieves the best RT-1 VM/VA performance and remains competitive on LIBERO and WidowX, suggesting that future latent supervision is particularly effective for modeling action outcomes and state transitions. MindLWPI obtains the best results on LIBERO and WidowX, achieves the strongest LIBERO-Plus result among the auxiliary-supervised paradigms, and remains close to the best RT-1 performance. On LIBERO-Plus, its total score of $74.8$ is also within $0.1$ points of the overall VLAFlow maximum achieved by MindPI (Frozen VLM). This indicates that language supervision and future latent supervision are complementary, with future latent alignment contributing strongly to cross-platform control transfer and language supervision improving overall robustness across benchmarks.

\subsection{Comparison with Public Baselines}

We report public baselines following the common format used in recent VLA technical reports. These baselines provide absolute-performance references, whereas VLAFlow internal variants are used for controlled comparison. Representative references include OpenVLA, OpenVLA-OFT, $\pi_0$, $\pi_{0.5}$, and GR00T-N1 \citep{kim2024openvla,kim2025openvlaoft,black2024pi0,black2025pi05,bjorck2025groot}.

\paragraph{LIBERO.}
Table~\ref{tab:libero_public_en} reports LIBERO results in terms of L-Spatial, L-Object, L-Goal, L-Long, and their average. Public baselines provide absolute-performance references, while VLAFlow variants are compared under the unified architecture and evaluation protocol.

\begin{table}[H]
\centering
\caption{Public baselines and VLAFlow results on the LIBERO benchmark (success rate, \%). External baselines are from public technical reports; VLAFlow variants are compared under a unified architecture and evaluation protocol.}
\label{tab:libero_public_en}
\begin{compacttable}[3.4pt]
\begin{adjustbox}{max width=0.92\textwidth}
\begin{tabular}{lccccc}
\toprule
Method & L-Spatial & L-Object & L-Goal & L-Long & Avg \\
\midrule
OpenVLA~\citep{kim2024openvla} & 84.7 & 88.4 & 79.2 & 53.7 & 76.5 \\
$\pi_0$-Fast~\citep{pertsch2025fast} & 96.4 & 96.8 & 88.6 & 60.2 & 85.5 \\
GROOT-N1~\citep{bjorck2025groot} & 94.4 & 97.6 & 93.0 & 90.6 & 93.9 \\
$\pi_0$~\citep{black2024pi0} & 98.0 & 96.8 & 94.4 & 88.4 & 94.4 \\
$\pi_{0.5}$~\cite{black2025pi05} & 98.8 & 98.2 & 98.0 & 92.4 & 96.9 \\
OpenVLA-OFT~\cite{kim2025openvlaoft} & 97.6 & 98.4 & 97.9 & 94.5 & 97.1 \\
\midrule
MindPI w/o PT & 97.8 & 99.0 & 97.0 & 94.0 & 97.0 \\
MindPI (Frozen VLM) & 98.4 & 99.2 & 96.8 & 94.6 & 97.2 \\
MindPI (Full PT) & 98.2 & 98.4 & 98.0 & 95.2 & 97.5 \\
MindLPI & 97.8 & 98.4 & 98.0 & 94.8 & 97.2 \\
MindWPI w/o PT & 98.0 & 99.6 & 98.2 & 93.6 & 97.4 \\
MindWPI & 99.0 & 99.6 & 98.6 & 96.8 & 98.5 \\
MindLWPI & \best{99.2} & \best{99.8} & \best{99.2} & \best{98.2} & \best{99.1} \\
\bottomrule
\end{tabular}
\end{adjustbox}
\end{compacttable}
\end{table}

VLAFlow reaches a level close to recent strong baselines on LIBERO. MindLWPI obtains high scores on all four suites, with a particularly large improvement on L-Long, suggesting that the combination of language supervision and future latent supervision helps long-horizon action chains. However, because LIBERO is nearly saturated overall, the following analysis focuses more on out-of-distribution transfer in LIBERO-Plus and SimplerEnv.

\paragraph{LIBERO-Plus.}
Table~\ref{tab:liberoplus_public_en} reports zero-shot robustness by official LIBERO-Plus perturbation category. Total is computed using the official aggregation protocol rather than a simple arithmetic mean over the seven perturbation types.

\begin{table}[H]
\centering
\caption{LIBERO-Plus zero-shot robustness comparison (success rate, \%). All methods are trained only on standard LIBERO and are evaluated on LIBERO-Plus without further adaptation. Total uses the official aggregation protocol rather than a simple average over the seven perturbations. Bold indicates the best result among VLAFlow variants in each column.}
\label{tab:liberoplus_public_en}
\scriptsize
\begin{adjustbox}{width=\textwidth}
\begin{tabular}{lcccccccc}
\toprule
Method & Camera & Robot & Language & Light & Background & Noise & Layout & Total \\
\midrule
OpenVLA~\citep{kim2024openvla} & 0.8 & 3.5 & 23.0 & 8.1 & 34.8 & 15.2 & 28.5 & 15.6 \\
OpenVLA-OFT~\citep{kim2025openvlaoft} & 56.4 & 31.9 & 79.5 & 88.7 & 93.3 & 75.8 & 74.2 & 69.6 \\
OpenVLA-OFT-w~\citep{kim2025openvlaoft} & 10.4 & 38.7 & 70.5 & 76.8 & 93.6 & 49.9 & 69.9 & 55.8 \\
OpenVLA-OFT-m~\citep{kim2025openvlaoft} & 55.6 & 21.7 & 81.0 & 92.7 & 91.0 & 78.6 & 68.7 & 67.9 \\
NORA~\citep{hung2025nora} & 2.2 & 37.0 & 65.1 & 45.7 & 58.6 & 12.8 & 62.1 & 39.0 \\
WorldVLA~\citep{cen2025worldvla} & 0.1 & 27.9 & 41.6 & 43.7 & 17.1 & 10.9 & 38.0 & 25.0 \\
UniVLA~\citep{bu2025univla} & 1.8 & 46.2 & 69.6 & 69.0 & 81.0 & 21.2 & 31.9 & 42.9 \\
$\pi_0$~\citep{black2024pi0} & 13.8 & 6.0 & 58.8 & 85.0 & 81.4 & 79.0 & 68.9 & 53.6 \\
$\pi_0$-Fast~\citep{pertsch2025fast} & 65.1 & 21.6 & 61.0 & 73.2 & 73.2 & 74.4 & 68.8 & 61.6 \\
RIPT-VLA~\cite{tan2025interactive} & 55.2 & 31.2 & 77.6 & 88.4 & 91.6 & 73.5 & 74.2 & 68.4 \\
\midrule
MindPI w/o PT & 26.1 & 32.0 & 71.0 & 88.0 & 94.4 & 58.3 & 70.3 & 59.9 \\
MindWPI w/o PT & 32.7 & 48.7 & 77.0 & 90.7 & 93.2 & 64.0 & 73.6 & 66.1 \\
MindPI (Frozen VLM) & 42.3 & 58.6 & 86.6 & 94.7 & 96.3 & \best{81.9} & 78.0 & \best{74.9} \\
MindPI (Full PT) & 42.0 & 57.8 & 75.4 & 91.5 & 93.2 & 57.4 & \best{86.7} & 68.8 \\
MindLPI & \best{46.2} & 61.7 & 84.6 & 93.4 & 95.2 & 58.4 & 82.2 & 72.3 \\
MindWPI & 36.7 & 73.8 & 86.2 & 96.5 & 91.6 & 54.2 & 84.8 & 72.6 \\
MindLWPI & 35.1 & \best{74.2} & \best{93.0} & \best{97.1} & \best{96.7} & 58.9 & 84.5 & 74.8 \\
\bottomrule
\end{tabular}
\end{adjustbox}
\end{table}

LIBERO-Plus reveals the different generalization preferences of different supervision signals under zero-shot perturbations. MindPI (Frozen VLM) achieves the highest VLAFlow total score and performs particularly well under sensor noise, indicating that freezing the VLM can preserve robust pretrained visual representations. MindPI (Full PT) performs best under object-layout perturbations, while MindLPI performs best under camera-viewpoint changes. Among the auxiliary-supervised paradigms, MindLWPI achieves the highest total score and leads on robot initial state, language, lighting, and background perturbations. Its total score is only $0.1$ points below MindPI (Frozen VLM), indicating that the combined language and future latent objectives provide broadly competitive zero-shot robustness without dominating every perturbation category.

\paragraph{SimplerEnv.}
Table~\ref{tab:simpler_public_en} reports SimplerEnv results in the format commonly used in public papers. Since the evaluation protocol is the same, we place VLAFlow variants alongside public baselines.

\begin{table}[H]
\centering
\caption{Public baselines and VLAFlow results on SimplerEnv (success rate, \%). RT-1 reports Visual Matching (VM) and Visual Augmentation (VA), while WidowX is the average over four tasks. VLAFlow uses Bridge-only fine-tuning for WidowX and RT-1-only fine-tuning for RT-1.}
\label{tab:simpler_public_en}
\begin{compacttable}[4pt]
\begin{adjustbox}{max width=0.9\textwidth}
\begin{tabular}{lcccc}
\toprule
Method & Size & RT-1 VM & RT-1 VA & WidowX \\
\midrule
SpatialVLA~\citep{qu2025spatialvla} & 4B & 75.1 & 70.7 & 42.7 \\
FPC-VLA~\citep{yang2026fpc} & 7B & 78.0 & 65.8 & 64.6 \\
MemoryVLA~\citep{shi2025memoryvla} & 7B & 77.7 & 72.7 & 71.9 \\
$\pi_0$~\citep{black2024pi0} & 3B & 58.8 & 56.8 & 27.8 \\
$\pi_0$+FAST~\citep{pertsch2025fast} & 3B & 61.9 & 60.5 & 39.5 \\
OpenVLA-OFT~\citep{kim2025openvlaoft} & 7B & 63.0 & 54.3 & 31.3 \\
DD-VLA~\citep{liang2025discrete} & 7B & 71.2 & 64.1 & 49.3 \\
\midrule
MindPI w/o PT & 4B & 75.7 & 60.4 & 59.6 \\
MindWPI w/o PT & 4B & 75.2 & 51.6 & 71.9 \\
MindPI (Frozen VLM) & 4B & 72.7 & 66.0 & 54.4 \\
MindPI (Full PT) & 4B & 68.2 & 55.5 & 65.9 \\
MindLPI & 4B & 74.6 & 59.2 & 65.6 \\
MindWPI & 4B & \best{86.7} & \best{71.1} & 74.5 \\
MindLWPI & 4B & 84.4 & 69.8 & \best{75.5} \\
\bottomrule
\end{tabular}
\end{adjustbox}
\end{compacttable}
\end{table}

SimplerEnv has a much larger distribution gap than LIBERO and therefore better exposes transfer differences among training objectives. Compared with mixed fine-tuning, platform-specific fine-tuning on Bridge and RT-1 better reflects whether a pre-trained representation can adapt to each target embodiment. MindWPI obtains the strongest RT-1 VM/VA results, showing that future latent alignment provides an effective training signal for modeling action outcomes and state transitions under cross-platform distribution shifts. In contrast, MindLWPI achieves the best WidowX performance and remains close to the best RT-1 scores. These results suggest that language supervision and future latent alignment are complementary, but their benefits are task- and platform-dependent: future latent alignment contributes more strongly to RT-1 transfer, whereas the combined language-and-latent supervision yields the most robust performance on WidowX.

\subsection{Mixed and Split Fine-tuning on SimplerEnv: Bias in a Low-Cost Ablation Environment}
\label{subsec:mixed_split}

In early hyperparameter ablations, we used a mixed SimplerEnv fine-tuning setting with both Bridge and RT-1 data to construct a lower-cost and faster-iteration downstream adaptation environment. This setting can provide rapid feedback on the relative trends of design choices such as future latent loss ratio, future-frame offset, and latent routing without requiring full per-platform fine-tuning. However, Bridge/WidowX and RT-1 differ in robot embodiment, visual distribution, action scale, and task composition; therefore, the absolute numbers from mixed fine-tuning may be affected by cross-platform data interference. We further compare mixed Simpler FT and Bridge-only FT on WidowX to calibrate the bias of this low-cost ablation environment.

\begin{table}[H]
\centering
\caption{WidowX comparison between mixed SimplerEnv fine-tuning and Bridge-only fine-tuning (success rate, \%). This table is only used to quantify the numerical bias of mixed fine-tuning as a low-cost hyperparameter-screening environment, and is not used to compare full pre-training gains.}
\label{tab:mixed_split_en}
\small
\begin{tabular}{lccc}
\toprule
Method & Mixed Simpler FT & Bridge-only FT & Change \\
\midrule
MindPI w/o PT & 63.0 & 59.6 & -3.4 \\
MindWPI w/o PT & 73.4 & 71.9 & -1.5 \\
MindPI (Full PT) & 55.5 & 65.9 & +10.4 \\
Action-only + current V-JEPA & 56.8 & 62.0 & +5.2 \\
MindWPI (Action-only FT) & 70.8 & 74.2 & +3.4 \\
MindWPI (Default) & 71.9 & 74.5 & +2.6 \\
\bottomrule
\end{tabular}
\end{table}

Table~\ref{tab:mixed_split_en} shows that mixed fine-tuning and Bridge-only fine-tuning exhibit a clear numerical discrepancy, and that the discrepancy varies across objective components. The matched current-context control changes from 56.8 to 62.0, whereas future-predictive pre-training with action-only fine-tuning changes from 70.8 to 74.2. For models without pre-training, mixed data may primarily increase data volume and act as regularization. For pre-trained models, however, the model has already acquired a certain cross-embodiment action prior from OXEMix; applying Bridge and RT-1 simultaneously during downstream adaptation may introduce cross-platform gradient conflicts or compromise optima, thereby weakening adaptation to a specific target platform. This interpretation is consistent with prior observations in multi-task gradient conflict and robot data-mixture optimization \citep{chen2018gradnorm,hejna2025remix}. Based on this observation, later ablations that use mixed Simpler FT are used only to compare relative trends among design choices. In the final SimplerEnv main results, we adopt a per-platform fine-tuning protocol: WidowX uses Bridge-only FT, and RT-1 uses RT-1-only FT.

\subsection{Ablation Studies}

\paragraph{Action--language correspondence in MindLPI.}
An alternative explanation for the gains of MindLPI is that the additional language-generation loss acts only as a generic auxiliary regularizer, without requiring correspondence between an action chunk and its verbal description. To test this explanation, we construct an in-batch language-target shuffling ablation. For a mini-batch $\mathcal{B}=\{(o_i,\ell_i,\mathbf{a}_i,y_i)\}_{i=1}^{B}$, where $y_i$ is the LAP-style description generated from action chunk $\mathbf{a}_i$, standard MindLPI uses the matched language target $y_i$. The shuffled variant samples a random permutation $\pi$ independently for each mini-batch and instead optimizes
\begin{equation}
\mathcal{L}_{\mathrm{shuffle}}
=\frac{1}{B}\sum_{i=1}^{B}\left[
\mathcal{L}_{\mathrm{act}}(o_i,\ell_i,\mathbf{a}_i)
+\lambda_{\mathrm{lang}}
\mathcal{L}_{\mathrm{lang}}(o_i,\ell_i,y_{\pi(i)})
\right].
\label{eq:mindlpi_shuffle}
\end{equation}
The observation, task instruction, continuous-action target, language-loss weight, and training protocol are otherwise unchanged. This intervention approximately preserves the vocabulary, template structure, sequence-length distribution, and mini-batch-level marginal distribution of the language targets, while disrupting their example-wise correspondence with the true actions.

\begin{table}[H]
\centering
\caption{MindLPI action--language correspondence ablation (success rate, \%). Language targets are randomly permuted within each pre-training mini-batch in the shuffled setting.}
\label{tab:mindlpi_shuffle_en}
\small
\begin{adjustbox}{max width=\textwidth}
\begin{tabular}{lccccc}
\toprule
Setting & LIBERO Avg & LIBERO-Plus Total & WidowX & RT-1 VM & RT-1 VA \\
\midrule
MindLPI & \best{97.2} & \best{72.3} & \best{65.6} & \best{74.6} & \best{59.2} \\
MindLPI w/ shuffled language targets & 97.0 & 65.8 & 47.1 & 73.2 & 33.3 \\
\midrule
Difference & $-0.2$ & $-6.5$ & $-18.5$ & $-1.4$ & $-25.9$ \\
\bottomrule
\end{tabular}
\end{adjustbox}
\end{table}

As shown in Table~\ref{tab:mindlpi_shuffle_en}, shuffling has little effect on the nearly saturated, in-distribution LIBERO benchmark, but substantially degrades transfer under distribution shift. In particular, success decreases by $6.5$ points on LIBERO-Plus, $18.5$ points on WidowX, and $25.9$ points on RT-1 VA; the reduction on RT-1 VM is smaller. These results are inconsistent with explaining MindLPI purely as generic regularization from an additional language loss. When the auxiliary objective and its marginal target distribution are retained but action--language correspondence is disrupted, transfer performance deteriorates markedly on several out-of-distribution evaluations. This supports the view that action-consistent language supervision provides a shared language-level representation of action intent across heterogeneous embodiments and action spaces. Nevertheless, shuffled targets also introduce contradictory supervision, so this experiment does not by itself establish fully interpretable semantic understanding. More conservatively, it demonstrates that example-wise action--language alignment is an important component of the MindLPI gains and cannot be replaced by an arbitrary auxiliary language-generation objective.

\paragraph{Gradient truncation in MindLPI.}
Table~\ref{tab:mindlpi_stopgrad_en} compares whether the action loss in MindLPI is backpropagated into the VLM. The initial motivation for gradient truncation was to prevent low-level action supervision from interfering with the VLM, but the results show that this strategy causes substantial degradation on LIBERO-Plus.

\begin{table}[H]
\centering
\caption{MindLPI gradient-truncation ablation (success rate, \%).}
\label{tab:mindlpi_stopgrad_en}
\small
\begin{tabular}{lcc}
\toprule
Method & LIBERO Avg & LIBERO-Plus Total \\
\midrule
MindLPI w/ stop-gradient & 96.3 & 45.8 \\
MindLPI w/o stop-gradient & \best{97.2} & \best{72.3} \\
\bottomrule
\end{tabular}
\end{table}

Removing gradient truncation is markedly better than using it, indicating that backpropagation from the action loss into the VLM is not merely noise; it may provide action-outcome alignment signals for vision-language representations. Therefore, the main MindLPI experiments use the no-stop-gradient setting.

\paragraph{Pre-training and downstream loss ratios in MindWPI.}
Table~\ref{tab:mindwpi_full_ablation_en} separates current-latent input, pre-training targets, and downstream fine-tuning loss ratios. The table reports both mixed SimplerEnv fine-tuning, used for low-cost screening, and Bridge-only fine-tuning, used to assess the WidowX target platform. All ratios are written as latent-to-action, namely $\lambda_{\mathrm{lat}}:\lambda_{\mathrm{act}}$, with the action weight normalized to one for joint objectives.

\begin{table}[H]
\centering
\caption{MindWPI objective ablation after OXEMix pre-training and SimplerEnv WidowX fine-tuning (success rate, \%). PT and FT ratios are $\lambda_{\mathrm{lat}}:\lambda_{\mathrm{act}}$; the action weight is normalized to one for joint objectives. Mixed FT is a low-cost screening protocol, while Bridge FT is the platform-specific result.}
\label{tab:mindwpi_full_ablation_en}
\scriptsize
\setlength{\tabcolsep}{2.8pt}
\begin{adjustbox}{max width=\textwidth}
\begin{tabular}{lcccrr}
\toprule
Method & PT target & PT lat.:act. & FT lat.:act. & Mixed FT & Bridge FT \\
\midrule
MindPI (Full PT) & action only & $0:1$ & $0:1$ & 55.5 & 65.9 \\
Action-only + current V-JEPA & action only & $0:1$ & $0:1$ & 56.8 & 62.0 \\
MindWPI (Latent-only PT) & latent only & $1:0$ & $0.1:1$ & 64.6 & 65.1 \\
MindWPI ($1:1$ FT) & act+lat & $1:1$ & $1:1$ & 61.5 & 70.6 \\
MindWPI (Action-only FT) & act+lat & $1:1$ & $0:1$ & 70.8 & 74.2 \\
MindWPI ($0.01:1$ FT) & act+lat & $1:1$ & $0.01:1$ & 67.2 & 66.4 \\
MindWPI ($0.1:1$ PT) & act+lat & $0.1:1$ & $0.1:1$ & 67.7 & 72.7 \\
MindWPI (Default) & act+lat & $1:1$ & $0.1:1$ & \best{71.9} & \best{74.5} \\
\bottomrule
\end{tabular}
\end{adjustbox}
\end{table}

The results show that future latent prediction is most effective when it is grounded by action supervision and balanced during pre-training. Latent-only pre-training reaches 64.6 under mixed fine-tuning and 65.1 under Bridge-only fine-tuning, compared with 55.5 and 65.9 for the matched MindPI full-pre-training baseline; it therefore does not provide a consistent transfer gain by itself. Under the same downstream ratio of $\lambda^{\mathrm{ft}}_{\mathrm{lat}}:\lambda^{\mathrm{ft}}_{\mathrm{act}}=0.1:1$, weakening the pre-training ratio to $0.1:1$ reaches 67.7/72.7, whereas balanced $1:1$ pre-training reaches 71.9/74.5 under mixed/Bridge adaptation.

The downstream fine-tuning ratio is also important. Starting from the same act+lat $(1:1)$ pre-trained checkpoint, a $1:1$ FT ratio performs poorly, while removing the latent loss improves transfer but remains below the recommended $0.1:1$ ratio. The matched ladder makes the component contributions explicit: adding current V-JEPA context changes the MindPI score by $+1.3$ points under mixed fine-tuning but $-3.9$ points under Bridge-only fine-tuning; enabling future prediction during pre-training then adds $+14.0$ and $+12.2$ points, respectively; retaining the weak downstream latent loss adds only $+1.1$ and $+0.3$ points. Thus, the dominant effect is predictive supervision during robot-data pre-training rather than access to current latent features alone.

\begin{table}[H]
\centering
\caption{Bridge-only task breakdown for the full-pre-training MindWPI ablations (success rate, \%). The aggregate scores correspond to Table~\ref{tab:mindwpi_full_ablation_en}.}
\label{tab:mindwpi_bridge_tasks_en}
\scriptsize
\setlength{\tabcolsep}{2.0pt}
\begin{adjustbox}{max width=\textwidth}
\begin{tabular}{llrrrrrrr}
\toprule
Setting & Protocol & Stack & Carrot & Spoon & Eggplant & Overall \\
\midrule
Action-only + current V-JEPA & Mixed & 21.9 & 47.9 & 71.9 & 85.4 & 56.8 \\
Action-only + current V-JEPA & Bridge & 19.8 & 61.5 & 67.7 & 99.0 & 62.0 \\
MindWPI (Latent-only PT) & Bridge & 18.8 & 55.2 & 86.5 & 100.0 & 65.1 \\
MindWPI ($1:1$ FT) & Bridge & 15.6 & 75.0 & 92.7 & 99.0 & 70.6 \\
MindWPI (Action-only FT) & Bridge & 35.4 & 65.6 & 96.9 & 99.0 & 74.2 \\
MindWPI ($0.01:1$ FT) & Bridge & 26.0 & 56.3 & 85.4 & 97.9 & 66.4 \\
MindWPI ($0.1:1$ PT) & Bridge & 53.1 & 61.5 & 83.3 & 92.7 & 72.7 \\
\bottomrule
\end{tabular}
\end{adjustbox}
\end{table}

The Bridge-only breakdown shows that future-predictive pre-training improves Stack, Carrot, and Spoon relative to the matched current-context control, while Eggplant is already near saturation. The aggregate gain is therefore not explained by the easiest task or by current latent tokens alone.

\paragraph{Low-cost w/o PT design ablations.}
Because full pre-training is expensive, we also conduct low-cost ablations of MindWPI design choices without robot-data pre-training. This group of experiments is not used to claim full pre-training gains; rather, it verifies whether future latent loss, future-frame offset, and latent routing design affect downstream adaptation.

\begin{table}[H]
\centering
\caption{MindWPI w/o PT design ablation (SimplerEnv WidowX success rate, \%). None of the rows uses robot-data pre-training.}
\label{tab:mindwpi_nopretrain_ablation_en}
\small
\begin{tabular}{lccc}
\toprule
Method & Loss setting & Future-frame offset & WidowX Avg \\
\midrule
MindWPI w/o PT (1:1 FT) & 1:1 & 8 & 64.3 \\
MindWPI w/o PT (Action-only FT) & action only & - & 69.0 \\
MindWPI w/o PT & 0.1:1 & 8 & \best{73.4} \\
MindWPI w/o PT (0.1:1, Offset 32) & 0.1:1 & 32 & 59.9 \\
\bottomrule
\end{tabular}
\end{table}

Without pre-training, downstream $\lambda^{\mathrm{ft}}_{\mathrm{lat}}:\lambda^{\mathrm{ft}}_{\mathrm{act}}=0.1:1$ remains significantly better than the $1:1$ setting, indicating that the latent loss ratio is also sensitive for downstream adaptation itself. Offset 32 is clearly weaker than the default Offset 8, suggesting that a future-frame interval that is too long may introduce state changes weakly associated with the current action.

\paragraph{MindLWPI and latent compression.}
Table~\ref{tab:compression_ablation_en} summarizes the results for latent-token compression and joint supervision in MindWPI/MindLWPI. AvgPool-k4 reduces the number of tokens from $256$ to $64$ while preserving strong performance; k16 and MLP compression are less stable in low-cost ablations. Therefore, the main text uses AvgPool-k4 as the default compression method for MindLWPI.

\begin{table}[H]
\centering
\caption{Latent compression and joint-supervision ablation (SimplerEnv WidowX success rate, \%). Compression experiments are mainly used to justify AvgPool-k4 as the default setting for MindLWPI.}
\label{tab:compression_ablation_en}
\small
\begin{tabular}{lccc}
\toprule
Setting & Compression method & Training stage & WidowX Avg \\
\midrule
MindWPI w/o PT & No compression & Direct downstream fine-tuning & 73.4 \\
MindWPI compressed & AvgPool-k4 & Direct downstream fine-tuning & 74.0 \\
MindWPI compressed & AvgPool-k16 & Direct downstream fine-tuning & 67.2 \\
MindWPI compressed & MLP-k4 & Direct downstream fine-tuning & 70.8 \\
MindWPI compressed & MLP-k16 & Direct downstream fine-tuning & 60.7 \\
MindLWPI & AvgPool-k4 & Full pre-training + Bridge FT & \best{75.5} \\
\bottomrule
\end{tabular}
\end{table}

\paragraph{Pre-training data composition.}
Table~\ref{tab:data_ablation_en} compares how different robot pre-training data sources affect the MindPI action-only modeling paradigm.

\begin{table}[H]
\centering
\caption{MindPI pre-training data-source ablation on SimplerEnv WidowX (success rate, \%). OXE raw subset denotes the original OXE subset including DROID, while excluding OXE-Augmented and RoboCOIN.}
\label{tab:data_ablation_en}
\small
\begin{tabular}{lccccc}
\toprule
Pre-training data & Stack & Carrot & Spoon & Eggplant & WidowX Avg \\
\midrule
No PT & 32.3 & 58.3 & 67.7 & \textbf{93.8} & 63.0 \\
RoboCOIN subset & 0.05 & 31.3 & 38.5 & 91.7 & 40.4 \\
OXE-Augmented & 13.5 & \textbf{59.4} & 84.4 & 81.3 & 59.7 \\
DROID only & \textbf{46.9} & 52.1 & 81.2 & 68.8 & 62.2 \\
OXE raw subset & 37.5 & 57.3 & \best{87.5} & 78.1 & \best{65.1} \\
\bottomrule
\end{tabular}
\end{table}

This ablation shows that the effect of action-only robot-data pre-training is highly dependent on the source and composition of the pre-training corpus. Different subsets lead to substantially different transfer behaviors: RoboCOIN subset causes severe degradation on WidowX, OXE-Augmented improves some manipulation categories but lowers the overall average, while DROID-only and OXE raw subset provide gains on certain tasks but still introduce task-specific regressions. These results suggest that negative transfer is not caused by robot-data pre-training per se, but by the difficulty of aligning heterogeneous embodiments, action definitions, task distributions, and visual domains under a purely low-dimensional action-supervision objective. This further supports our view that large-scale heterogeneous VLA pre-training requires intermediate constraints beyond action imitation alone in order to form stable and transferable action representations.

\subsection{Future-Latent Fidelity and Closed-Loop Execution}
\label{subsec:future_alignment_mechanism}

The preceding results show that future latent alignment improves downstream transfer, but success-rate gains alone do not determine whether the future latent objective acts merely as a generic auxiliary regularizer or learns state-transition representations that remain relevant to closed-loop execution. To investigate this mechanism, we examine whether future latent prediction fidelity is associated with the realized execution outcome, and whether this relationship persists when the future latent objective is removed during downstream fine-tuning.

\paragraph{Trajectory-level evaluation protocol.}
We collect closed-loop trajectories from four SimplerEnv WidowX tasks: putting a carrot on a plate, putting a spoon on a towel, stacking two blocks, and putting an eggplant into a basket. For each model and task, we record $20$ episodes using the original evaluation pipeline, resulting in $80$ trajectories with episode-level success labels per model. The recorder passively stores the observations, executed actions, and outcomes without modifying policy inference, action sampling, or environment interaction; the latent-quality analysis itself uses only the observation sequence and outcome label. We then replay the recorded observations offline and reproduce the latent-prediction computation used during training, without providing a ground-truth future frame to the predictor or resampling policy actions.

Let $o_t$ denote the visual observation at control step $t$, $\ell$ the language instruction, and $E_{\mathrm{VJEPA}}$ the frozen V-JEPA~2 feature extractor. The current latent and the realized future latent are
\begin{equation}
\mathbf{Z}_t=E_{\mathrm{VJEPA}}(o_t),
\qquad
\mathbf{Z}_{t+\Delta}=E_{\mathrm{VJEPA}}(o_{t+\Delta}),
\qquad \Delta=8.
\end{equation}
The prediction horizon follows the training configuration. Given the current VLM context $\mathbf{H}^{\mathrm{VLM}}_t$ and current visual latent, the model predicts
\begin{equation}
\widehat{\mathbf{Z}}_{t+\Delta}
=
g_\theta\!\left(
\mathbf{H}^{\mathrm{VLM}}_t,\mathbf{Z}_t
\right).
\end{equation}
The structured attention mask prevents latent tokens from attending to action tokens. Consequently, the recovered latent prediction depends only on the current vision-language context and current latent representation, rather than on the realized future actions or the sampled action sequence.

For MindLWPI with AvgPool-k4, prediction quality is evaluated in the same compressed space used by the training objective. Every four adjacent V-JEPA tokens are averaged:
\begin{equation}
\widetilde{\mathbf{z}}_{t,m}
=
\frac{1}{4}
\sum_{j=1}^{4}
\mathbf{z}_{t,4(m-1)+j}.
\end{equation}
The same operation is applied to the current latent and realized future target, ensuring that both lie in the same $64$-token space. Because MindWPI and MindLWPI use different latent tokenizations and comparison spaces, we compare successful and failed trajectories within each model and do not rank the models by their absolute error values.

\paragraph{Metrics and statistical unit.}
For the $n$-th predicted and target latent tokens, we use token-wise cosine distance as the primary metric:
\begin{equation}
e_{\mathrm{cos}}(t)
=
1-\frac{1}{N}
\sum_{n=1}^{N}
\frac{
\widehat{\mathbf{z}}_{t+\Delta,n}^{\top}
\mathbf{z}_{t+\Delta,n}
}{
\left\|\widehat{\mathbf{z}}_{t+\Delta,n}\right\|_2
\left\|\mathbf{z}_{t+\Delta,n}\right\|_2
}.
\end{equation}
For episode $i$, we average over all temporally aligned control steps:
\begin{equation}
\overline e_i
=
\frac{1}{T_i-\Delta}
\sum_{t=1}^{T_i-\Delta}e_i(t).
\end{equation}
All significance tests therefore treat an episode, rather than an individual control step, as the statistical unit, avoiding the assumption that temporally correlated observations from the same trajectory are independent. We compare successful and failed episodes using a two-sided Mann--Whitney $U$ test and report Cohen's $d$ as the effect size. Element-wise MSE is additionally used as a complementary metric.

\begin{table}[H]
\centering
\caption{\textbf{Future-latent prediction error versus closed-loop execution outcome.} Error S/F denotes the mean episode-level token-wise cosine distance over successful and failed executions. A positive Cohen's $d$ indicates higher prediction error for failed trajectories. MindLWPI is evaluated in its AvgPool-k4 compressed space, so its absolute error is not directly comparable with that of uncompressed MindWPI.}
\label{tab:future_alignment_outcome}
\small
\setlength{\tabcolsep}{4.5pt}
\begin{tabular}{lcccccc}
\toprule
Model & FT ratio $\lambda^{\mathrm{ft}}_{\mathrm{lat}}:\lambda^{\mathrm{ft}}_{\mathrm{act}}$
& S/F & Error S & Error F & Cohen's $d$ & $p$ \\
\midrule
MindWPI & $0:1$   & $61/19$ & 0.3491 & 0.3541 & 0.22 & 0.830 \\
MindWPI & $0.1:1$ & $61/19$ & 0.2305 & 0.2917 & 1.42 & $1.31{\times}10^{-6}$ \\
MindLWPI (AvgPool-k4) & $0.1:1$ & $65/15$ & 0.1056 & 0.1477 & 2.44 & $7.17{\times}10^{-8}$ \\
\bottomrule
\end{tabular}
\end{table}

\begin{figure}[H]
\centering
\setlength{\tabcolsep}{2pt}
\begin{tabular}{ccc}
\textbf{MindWPI, FT $0:1$} &
\textbf{MindWPI, FT $0.1:1$} &
\textbf{MindLWPI, FT $0.1:1$} \\
\includegraphics[width=0.32\textwidth]{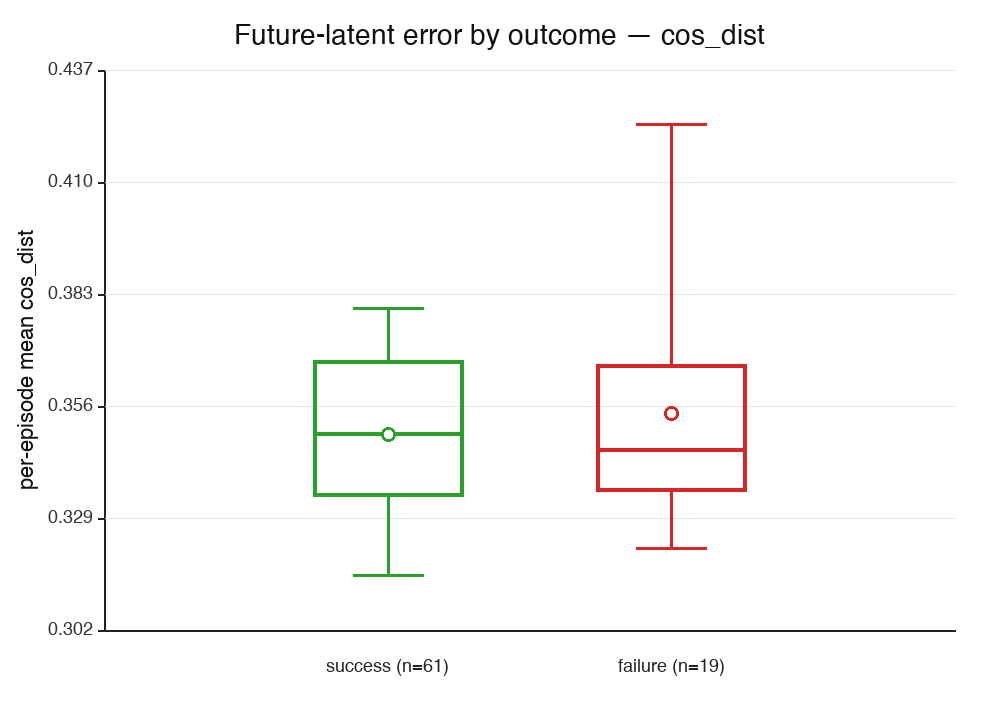} &
\includegraphics[width=0.32\textwidth]{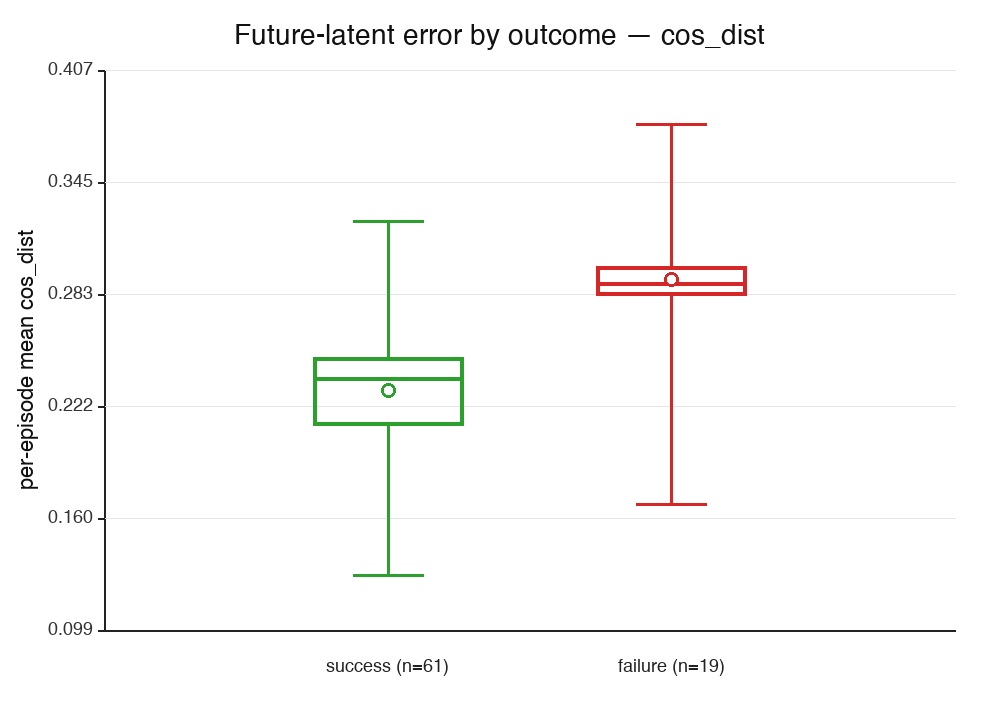} &
\includegraphics[width=0.32\textwidth]{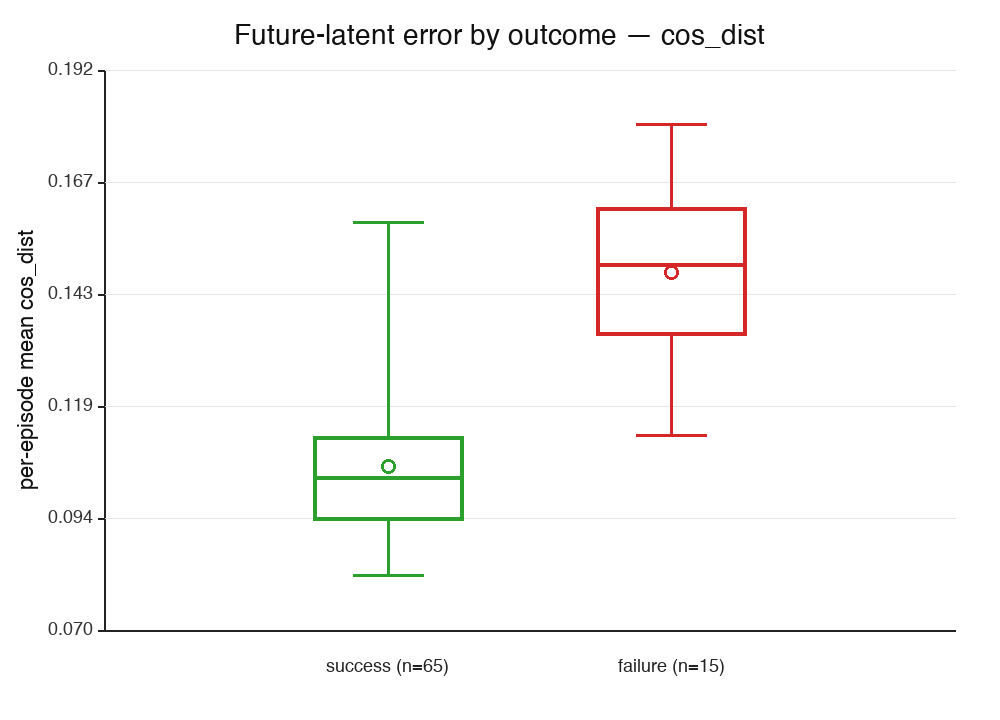} \\
\includegraphics[width=0.32\textwidth]{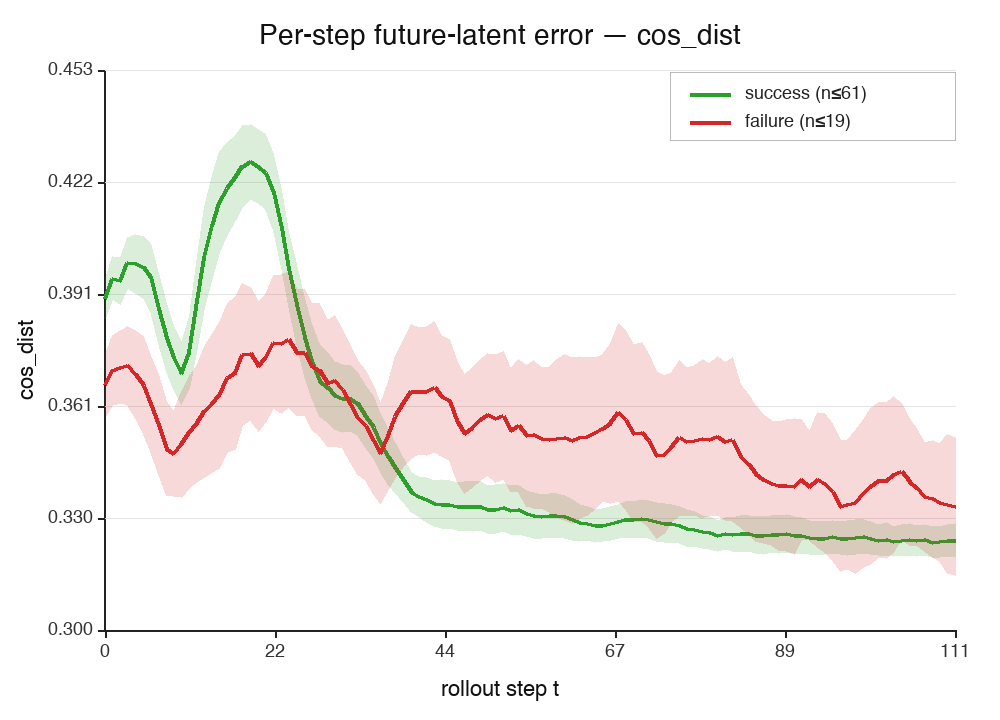} &
\includegraphics[width=0.32\textwidth]{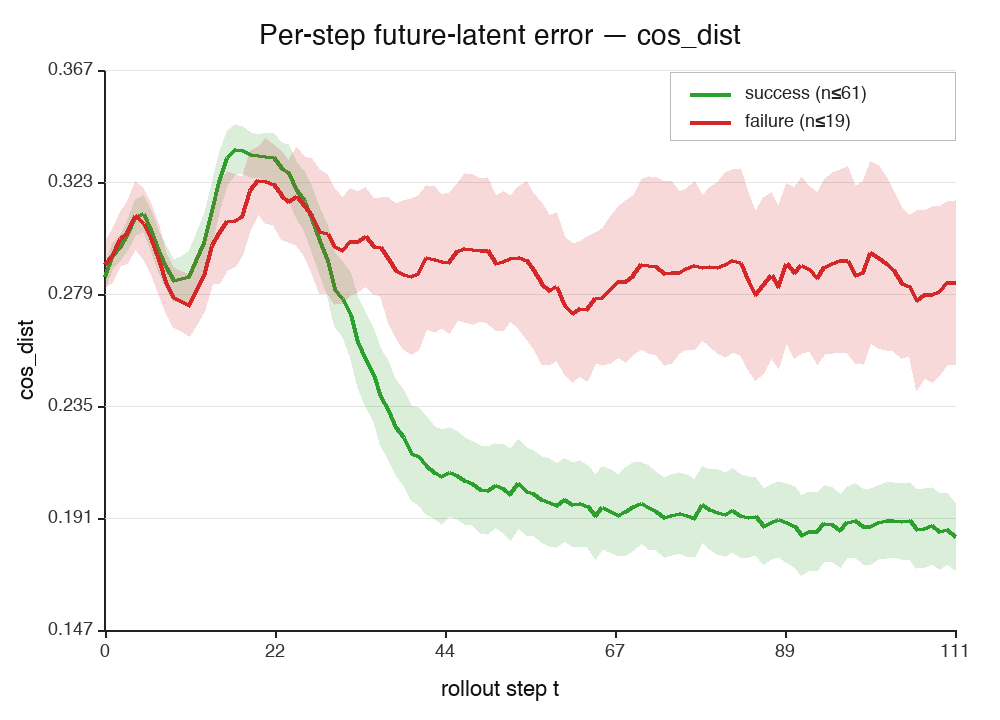} &
\includegraphics[width=0.32\textwidth]{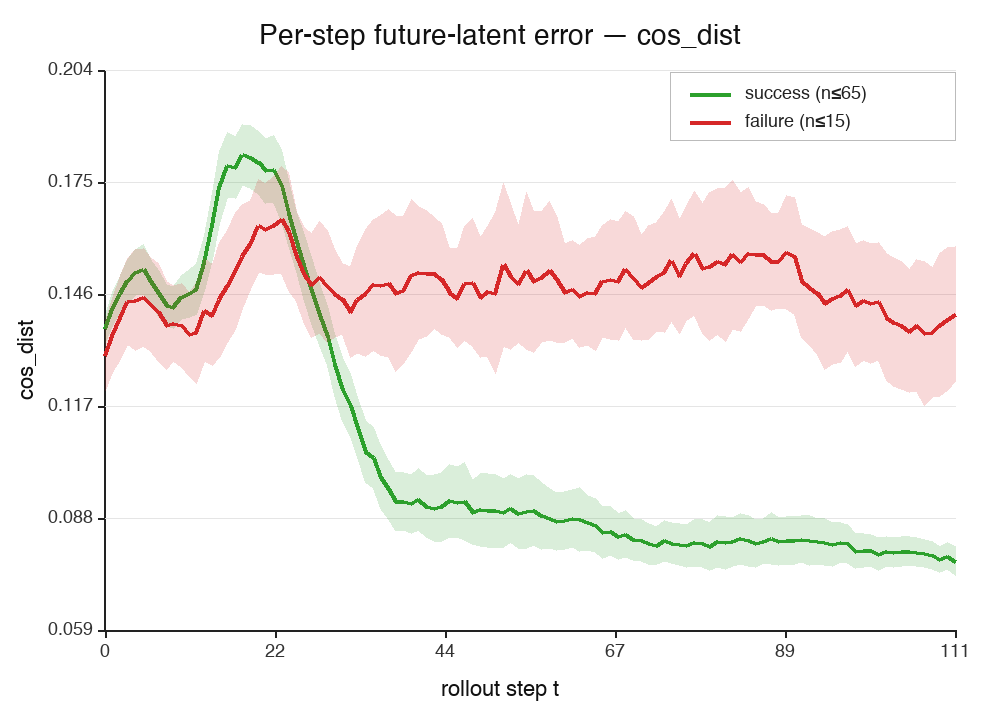}
\end{tabular}
\caption{\textbf{Future-latent fidelity and closed-loop execution.} The top row compares episode-level mean token-wise cosine distance between successful and failed executions; the bottom row shows the mean per-step error with $95\%$ confidence intervals. When downstream latent alignment is retained, successful trajectories have significantly lower aggregate prediction error and progressively separate from failed trajectories during execution. This relationship disappears when the downstream latent objective is disabled. MindLWPI errors are measured in its compressed representation space and are not directly comparable in absolute magnitude with MindWPI errors.}
\label{fig:future_alignment_outcome}
\end{figure}

\paragraph{Prediction fidelity tracks successful execution.}
As shown in Table~\ref{tab:future_alignment_outcome}, when downstream fine-tuning retains a weak future latent objective with $\lambda^{\mathrm{ft}}_{\mathrm{lat}}:\lambda^{\mathrm{ft}}_{\mathrm{act}}=0.1:1$, successful MindWPI trajectories have substantially lower prediction error than failed trajectories. Their mean cosine distances are $0.2305$ and $0.2917$, respectively, corresponding to a large effect size of $d=1.42$ and a statistically significant difference of $p=1.31{\times}10^{-6}$. Episode-level cosine error is negatively correlated with success, with Pearson $r=-0.535$ and Spearman $\rho=-0.545$. MSE gives the same conclusion: successful and failed trajectories have mean errors of $3.5239$ and $4.2928$, respectively, with $d=1.26$ and $p=9.81{\times}10^{-7}$.

In contrast, when the downstream latent objective is disabled, the mean cosine distances of successful and failed MindWPI trajectories are $0.3491$ and $0.3541$, and the difference is not significant ($d=0.22$, $p=0.830$). The corresponding MSE comparison is also non-significant. Thus, the association between future-prediction fidelity and execution outcome does not arise automatically from the MindWPI architecture; it emerges when future latent alignment is retained during downstream adaptation.

MindLWPI provides supporting evidence that the same relationship persists under a different latent tokenization. Across its $80$ trajectories, the mean cosine distances of successful and failed episodes are $0.1056$ and $0.1477$, respectively, yielding $d=2.44$ and $p=7.17{\times}10^{-8}$. Episode-level cosine error has Pearson $r=-0.711$ and Spearman $\rho=-0.607$ with success. MSE again yields a significant separation: $1.0287$ for successful trajectories versus $1.4060$ for failed trajectories, with $d=2.38$ and $p=9.43{\times}10^{-8}$. Although the absolute errors of the compressed and uncompressed models are not directly comparable, the consistent within-model separation indicates that the relationship between future latent fidelity and execution outcome is preserved across the two formulations.

\paragraph{Temporal pattern and interpretation.}
The per-step curves in Figure~\ref{fig:future_alignment_outcome} reveal a more detailed temporal pattern. With downstream latent alignment, successful and failed trajectories have similar errors during the initial stage, but their curves progressively separate after approximately $30$ control steps. Prediction error continues to decrease along successful trajectories, whereas it remains relatively high along failed trajectories. MindWPI without downstream latent loss does not exhibit a stable late-stage separation. The first $10$ aligned control steps likewise show no significant separation in the expected direction. Future latent error is therefore better interpreted as an online trajectory-consistency signal than as an early predictor that determines success from the initial observation.

This analysis establishes a strong trajectory-level association rather than a complete proof of a unidirectional causal relationship. The realized future observation $o_{t+\Delta}$ is jointly produced by the evaluated policy actions and the environment response. Lower error may therefore indicate that future-state modeling supplies a more effective dynamics constraint for action generation, but it may also partially reflect that successful, in-distribution trajectories are intrinsically easier to predict. These explanations are not mutually exclusive. Nevertheless, the success--failure separation disappears when the downstream latent objective is disabled and emerges consistently when it is retained in both MindWPI and MindLWPI. This evidence suggests that future latent alignment is not merely a control-agnostic auxiliary regularizer; it induces a predictive representation whose fidelity tracks closed-loop trajectory quality, providing mechanism-level evidence for its contribution to state-transition modeling and action execution.

\subsection{Efficient Adaptation with LoRA}

\begin{figure}[H]
\centering
\includegraphics[width=0.62\textwidth]{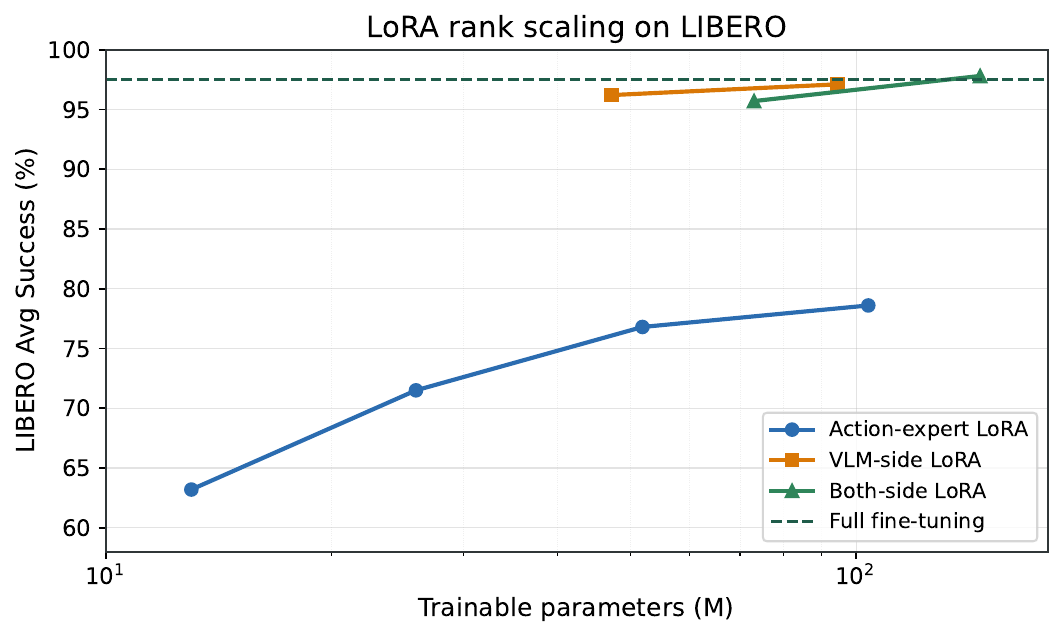}
\caption{LoRA rank scaling on LIBERO. The horizontal axis is the number of trainable parameters (M), and the vertical axis is the average success rate over the four LIBERO suites. The horizontal dashed line indicates the MindPI (Full PT) full-parameter fine-tuning baseline.}
\label{fig:lora_rank_en}
\end{figure}

Considering the storage and computational cost of full-parameter fine-tuning, we additionally evaluate low-rank adaptation with LoRA \citep{hu2021lora}. This experiment inherits the OXEMix pre-training weights from MindPI (Full PT) and is tested only on LIBERO for rapid evaluation; it is not a core variable in the comparison of training paradigms. Figure~\ref{fig:lora_rank_en} shows LIBERO average success rate as the number of trainable parameters varies across LoRA injection locations, and Table~\ref{tab:lora_raw_en} reports complete results for the four LIBERO suites.

\begin{table}[H]
\centering
\caption{Efficient LoRA adaptation results on LIBERO (success rate, \%). LIBERO contains four suites: L-Spatial, L-Object, L-Goal, and L-Long. Avg is the average over the four suites.}
\label{tab:lora_raw_en}
\scriptsize
\begin{adjustbox}{width=\textwidth}
\begin{tabular}{lccccc c}
\toprule
Method & Trainable parameters & L-Spatial & L-Object & L-Goal & L-Long & Avg \\
\midrule
Action LoRA r=64 & 13.0M & 71.8 & 85.2 & 55.2 & 40.6 & 63.2 \\
Action LoRA r=128 & 25.9M & 83.0 & 88.8 & 64.8 & 49.4 & 71.5 \\
Action LoRA r=256 & 51.9M & 88.4 & 93.6 & 69.2 & 56.0 & 76.8 \\
Action LoRA r=512 & 103.8M & 87.8 & 95.0 & 68.8 & 63.0 & 78.6 \\
VLM LoRA r=128 & 47.2M & 97.4 & 95.8 & 99.2 & 92.2 & 96.2 \\
VLM LoRA r=256 & 94.4M & 98.2 & 99.2 & 98.8 & 92.2 & 97.1 \\
Both LoRA r=128 & 73.1M & 96.8 & 99.0 & 97.8 & 89.2 & 95.7 \\
Both LoRA r=256 & 146.3M & 98.0 & 99.6 & 98.8 & 94.8 & \best{97.8} \\
\bottomrule
\end{tabular}
\end{adjustbox}
\end{table}

The results show that injecting LoRA only into the action expert makes it difficult to approach full-parameter fine-tuning. VLM-side LoRA already approaches full fine-tuning with around $100$M trainable parameters, while both-side LoRA achieves the highest average score under a larger parameter budget. This suggests that for VLM-based VLA models, downstream adaptation requires not only updating the action expert, but also providing sufficient low-rank adaptation capacity for vision-language representations.

\subsection{Meta-Action Hypothesis and Representation Analysis}
\label{subsec:meta_action}

\begin{figure}[H]
\centering
\includegraphics[width=0.76\textwidth]{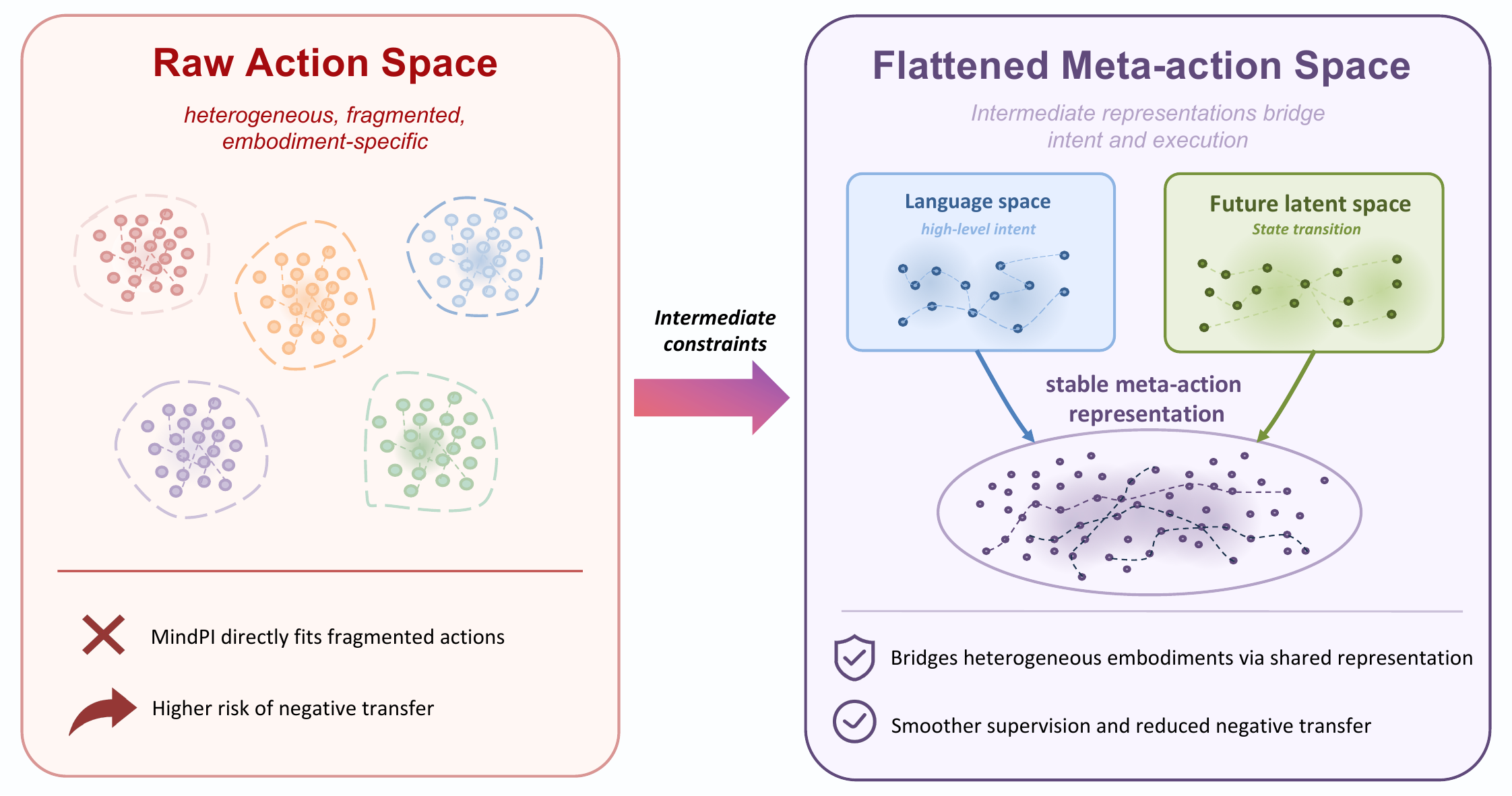}
\caption{Meta-action hypothesis for heterogeneous VLA pre-training. MindPI directly fits heterogeneous action labels and is therefore exposed to differences in embodiment, control frequency, and action definition. MindLPI and MindWPI introduce intermediate constraints through language-level action intent and future visual state transitions, respectively. MindLWPI combines the two signals. We hypothesize that these intermediate constraints reduce source-specific fragmentation while preserving action-relevant information.}
\label{fig:meta_action_en}
\end{figure}

We operationalize a \emph{meta-action representation} as one whose geometry is less dominated by the source dataset while retaining the context required for action generation. This definition does not require complete domain invariance; rather, samples from heterogeneous sources should exhibit less source-specific fragmentation in the action representation space. We next examine this prediction by comparing intermediate representations from MindPI, MindWPI, and MindLWPI.

\paragraph{Analysis protocol.}
We construct a balanced analysis set by sampling $50$ examples from each of $25$ raw OXE subsets, yielding $1{,}250$ examples in total. A fixed sample manifest is shared across all models. For each example, we run seeded four-step action denoising with checkpoints selected at approximately matched pre-training exposure, and capture intermediate representations using forward hooks without modifying the model inference path. We analyze three middle-to-late depths,
$\mathcal{D}=\{18,20,22\}$, from both the VLM and the DiT action expert.

Let $\mathbf{H}^{\mathrm{vlm}}_l\in\mathbb{R}^{T_v\times d_v}$ denote the VLM hidden states at depth $l$, and let $\mathbf{m}\in\{0,1\}^{T_v}$ denote the valid-token mask. We obtain one VLM feature vector per example by masked mean pooling over all valid vision-language tokens:
\begin{equation}
\mathbf{z}^{\mathrm{vlm}}_l
=
\frac{\sum_{i=1}^{T_v}m_i\mathbf{H}^{\mathrm{vlm}}_{l,i}}
{\max\left(1,\sum_{i=1}^{T_v}m_i\right)}.
\label{eq:vlm_rep_pool}
\end{equation}
For MindWPI and MindLWPI, the state and current visual latent tokens are first processed by a prefix-prefill pass, and their layer-wise key-value representations are cached. Each subsequent denoising pass processes only the $T_a=16$ action tokens, which attend to the VLM context and the cached latent prefix. Consequently, the captured DiT outputs contain the same number and type of action-token positions for all three models. Let
$\mathbf{H}^{\mathrm{act}}_{l,S}\in\mathbb{R}^{T_a\times d_a}$
be the action-token hidden states at depth $l$ in the final denoising iteration, where $S=4$. The corresponding DiT feature is
\begin{equation}
\mathbf{z}^{\mathrm{dit}}_l
=
\frac{1}{T_a}\sum_{i=1}^{T_a}\mathbf{H}^{\mathrm{act}}_{l,S,i}.
\label{eq:dit_rep_pool}
\end{equation}
Thus, differences in the DiT visualizations reflect how the respective training objectives and contextual signals reorganize action-token representations, rather than a direct averaging effect from different numbers of latent tokens.

For every model and depth, the resulting $1{,}250$ feature vectors are independently projected to two dimensions using t-SNE with perplexity $30$, PCA initialization, and random seed $0$. Points are colored by their source datasets. We additionally apply HDBSCAN to each two-dimensional embedding with minimum cluster size $45$ and minimum samples $22$; red dashed enclosures indicate the detected communities. Since t-SNE and HDBSCAN can depend on projection and clustering hyperparameters, we interpret these visualizations as qualitative evidence and focus on patterns that remain consistent across the three analyzed depths.

\paragraph{Representation-level evidence.}
MindPI exhibits pronounced dataset-specific islands in both representation spaces. Across VLM depths $18$, $20$, and $22$, it forms $20$ HDBSCAN communities, while its DiT action-token representations form $21$, $20$, and $21$ communities, respectively. Future latent alignment has a distinctly localized effect. The VLM representations of MindWPI remain strongly organized by source dataset, forming $22$--$23$ communities, whereas its DiT representations consistently form only $14$ communities. This reduction is consistent across all three depths. The pattern suggests that future latent supervision primarily reorganizes the action-generation space rather than globally suppressing source structure in the VLM.

MindLWPI preserves this cross-dataset consolidation in the action expert, producing $13$--$14$ DiT communities across the three depths. Its VLM embeddings also form fewer detected communities ($10$--$12$) than those of MindPI and MindWPI, with several communities containing samples from multiple datasets. Because the comparison is based on two-dimensional projections, we interpret this pattern as a change in projected representation geometry rather than evidence that dataset information has been eliminated. The observed effect should therefore be understood as \emph{partial cross-dataset consolidation}, rather than complete domain invariance.

Taken together, the visual analysis provides qualitative representation-level evidence consistent with the meta-action hypothesis. Future latent prediction supplies a shared state-transition reference that reduces source-specific fragmentation among action-token representations across multiple DiT depths.

\begin{figure}[p]
\centering
\includegraphics[width=\textwidth]{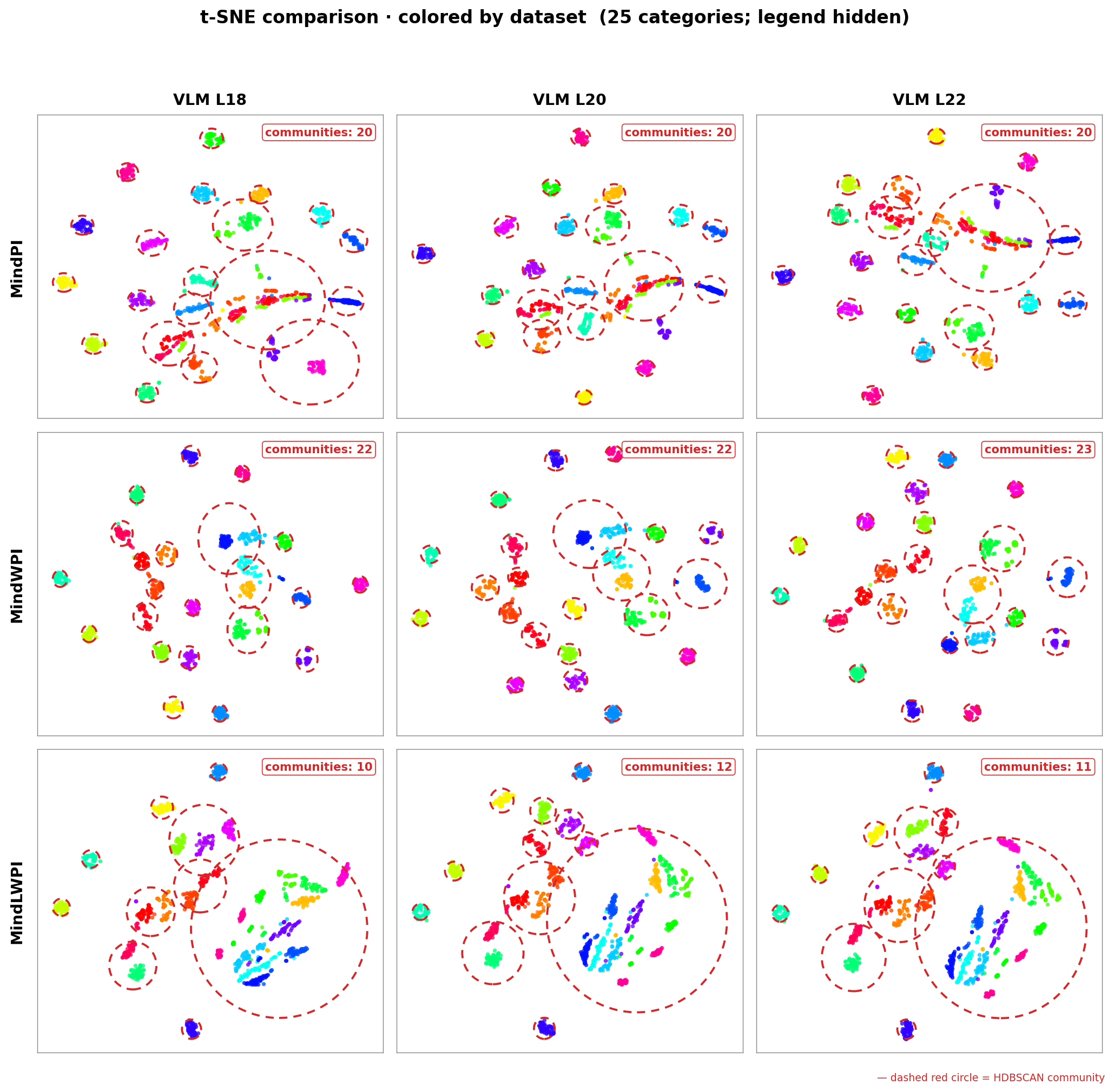}
\caption{\textbf{VLM representation geometry across source datasets.} Each row corresponds to a pre-training paradigm and each column to a VLM depth. Every point represents one of the $1{,}250$ shared OXE samples and is colored by its source dataset. Red dashed enclosures show HDBSCAN communities detected in the two-dimensional t-SNE embedding. MindWPI remains strongly source-structured, whereas the projected MindLWPI representations form fewer detected communities across the three depths.}
\label{fig:tsne_vlm_dataset}
\end{figure}

\begin{figure}[p]
\centering
\includegraphics[width=\textwidth]{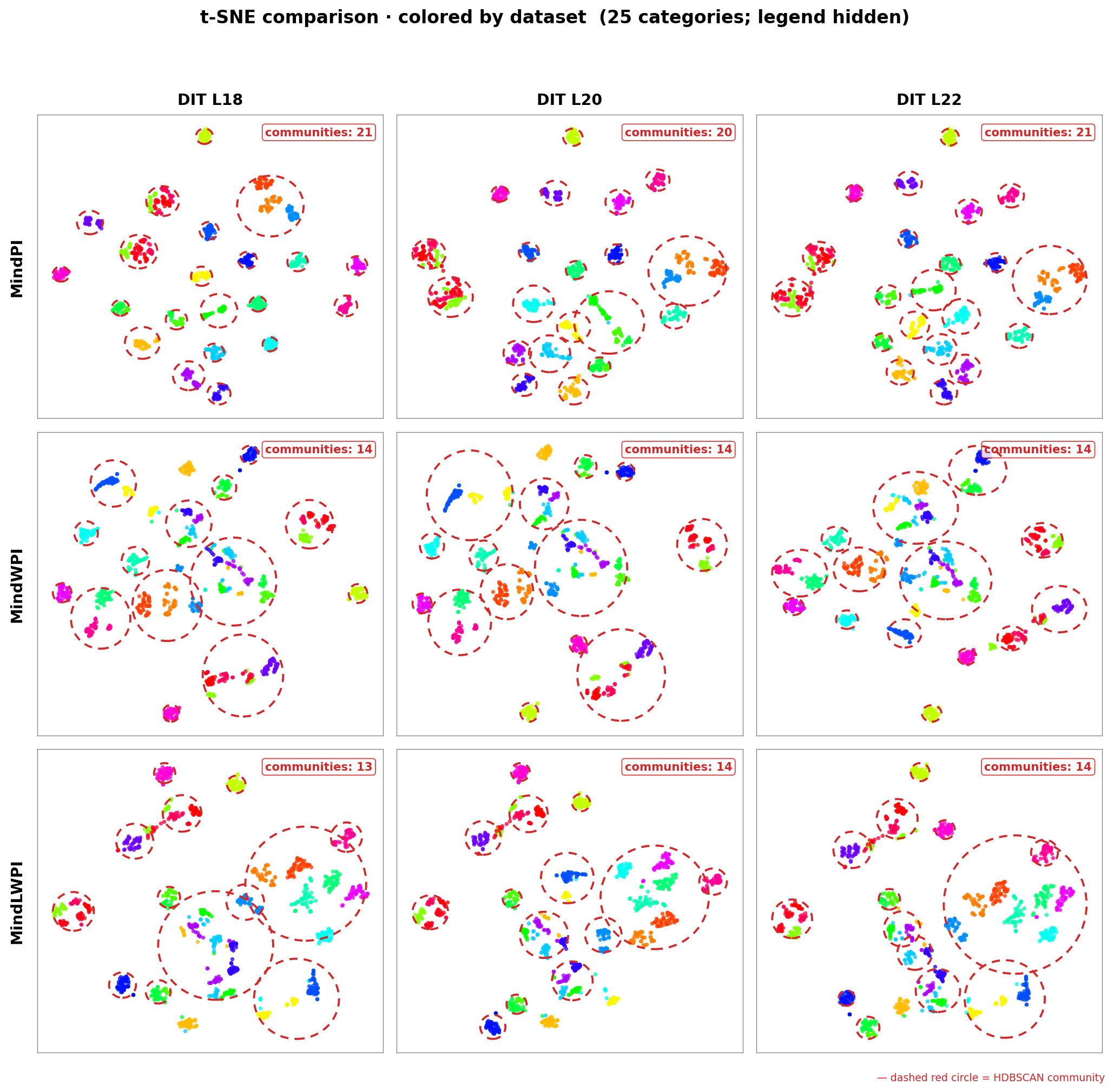}
\caption{\textbf{Action-token representation geometry across source datasets.} The plotted feature at each depth is obtained by mean pooling the $16$ action-token hidden states from the final denoising iteration. MindPI exhibits pronounced dataset-specific islands, whereas MindWPI and MindLWPI form fewer, more cross-dataset communities consistently across the three displayed depths. Colors and community enclosures follow Figure~\ref{fig:tsne_vlm_dataset}.}
\label{fig:tsne_dit_dataset}
\end{figure}

\FloatBarrier

\section{Conclusion}

This report presents VLAFlow, a unified flow-matching framework for controlled comparison of VLA training paradigms. Using approximately $5{,}000$ hours of OXEMix heterogeneous robot data, a unified $\pi_0$-style architecture, and a unified $14$-dimensional action space, we compare four paradigms: action-only modeling (MindPI), language-supervised co-training (MindLPI), future latent alignment (MindWPI), and their combination (MindLWPI). Experiments show that action-only pre-training is sensitive to data distribution and VLM update strategy, leading to unstable transfer under large distribution gaps. Language supervision and future latent supervision provide complementary intermediate constraints from action intent and state transition, respectively. MindWPI achieves the strongest RT-1 transfer performance, highlighting the value of future latent alignment for cross-platform control. MindLWPI further combines language and future latent supervision, obtaining the best results on LIBERO and WidowX, the strongest zero-shot LIBERO-Plus result among auxiliary-supervised paradigms, and competitive performance on RT-1. The t-SNE feature analysis further shows that future latent alignment consistently reduces source-specific fragmentation in action-token representations across multiple DiT depths, providing qualitative representation-level evidence for the proposed meta-action-space view. Future work will scale pre-training, explore pre-training-stage loss ratios, and validate the framework on real robot platforms.

\bibliographystyle{plainnat}
\bibliography{references}

\appendix
\section{Implementation Details}
\label{app:implementation}

This appendix supplements the implementation details omitted from Section~\ref{subsec:shared_architecture}. The main text retains only details directly related to the paradigm comparison, while this section supports reproducibility.

\subsection{VLM and Action Expert Configuration}
\label{app:model_config}

All models use Qwen3-VL-4B-Instruct as the vision-language backbone \citep{bai2025qwen3vl}. Its language backbone is a Transformer with $L=36$ layers, hidden dimension $d_{\mathrm{vlm}}=2048$, and $H=16$ attention heads. It uses grouped-query attention (GQA) with $H_{\mathrm{kv}}=8$ key-value heads and head dimension $d_h=128$. The action expert consists of $N=36$ DiT blocks with internal hidden dimension $d_a=1280$. To support concatenation with VLM caches, the key-value head count and head dimension of the action expert are aligned with those of the VLM.

The action expert input is a sequence of noised action tokens with default length $T=16$. The framework retains interfaces for robot-state tokens and learnable context tokens, but they are not enabled in the experiments reported here. MindWPI and MindLWPI additionally concatenate current-frame latent-space tokens as a prefix to the action expert.

\subsection{Key-Value Cache Sharing}
\label{app:kv_cache}

During the VLM forward pass, we store the key-value cache at every layer, $\{(\mathbf{K}^{\mathrm{vlm}}_l,\mathbf{V}^{\mathrm{vlm}}_l)\}_{l=1}^{L}$, where
\begin{equation}
\mathbf{K}^{\mathrm{vlm}}_l,\mathbf{V}^{\mathrm{vlm}}_l\in\mathbb{R}^{B\times H_{\mathrm{kv}}\times S_{\mathrm{vlm}}\times d_h}.
\end{equation}
At action-expert layer $i$, queries, keys, and values $\mathbf{Q}_i,\mathbf{K}_i,\mathbf{V}_i$ are produced from the action-expert tokens. The VLM cache at layer $i$ is concatenated with the action expert's own keys and values:
\begin{equation}
\tilde{\mathbf{K}}_i=[\mathbf{K}^{\mathrm{vlm}}_i;\mathbf{K}_i],\qquad
\tilde{\mathbf{V}}_i=[\mathbf{V}^{\mathrm{vlm}}_i;\mathbf{V}_i].
\end{equation}
The attention at this layer is
\begin{equation}
\Attnop(\mathbf{Q}_i,\tilde{\mathbf{K}}_i,\tilde{\mathbf{V}}_i)
=\softmaxop\left(\frac{\mathbf{Q}_i\tilde{\mathbf{K}}_i^{\top}}{\sqrt{d_h}}+\mathbf{M}\right)\tilde{\mathbf{V}}_i.
\end{equation}
Action-expert layers and VLM layers are aligned layer by layer by default, although a layer-offset parameter can also be used. At inference time, the VLM cache and static prefix cache can be reused across denoising steps to reduce control latency.

\subsection{Attention Mask}
\label{app:mask}
\begin{figure}[H]
    \centering
    \includegraphics[width=0.86\textwidth]{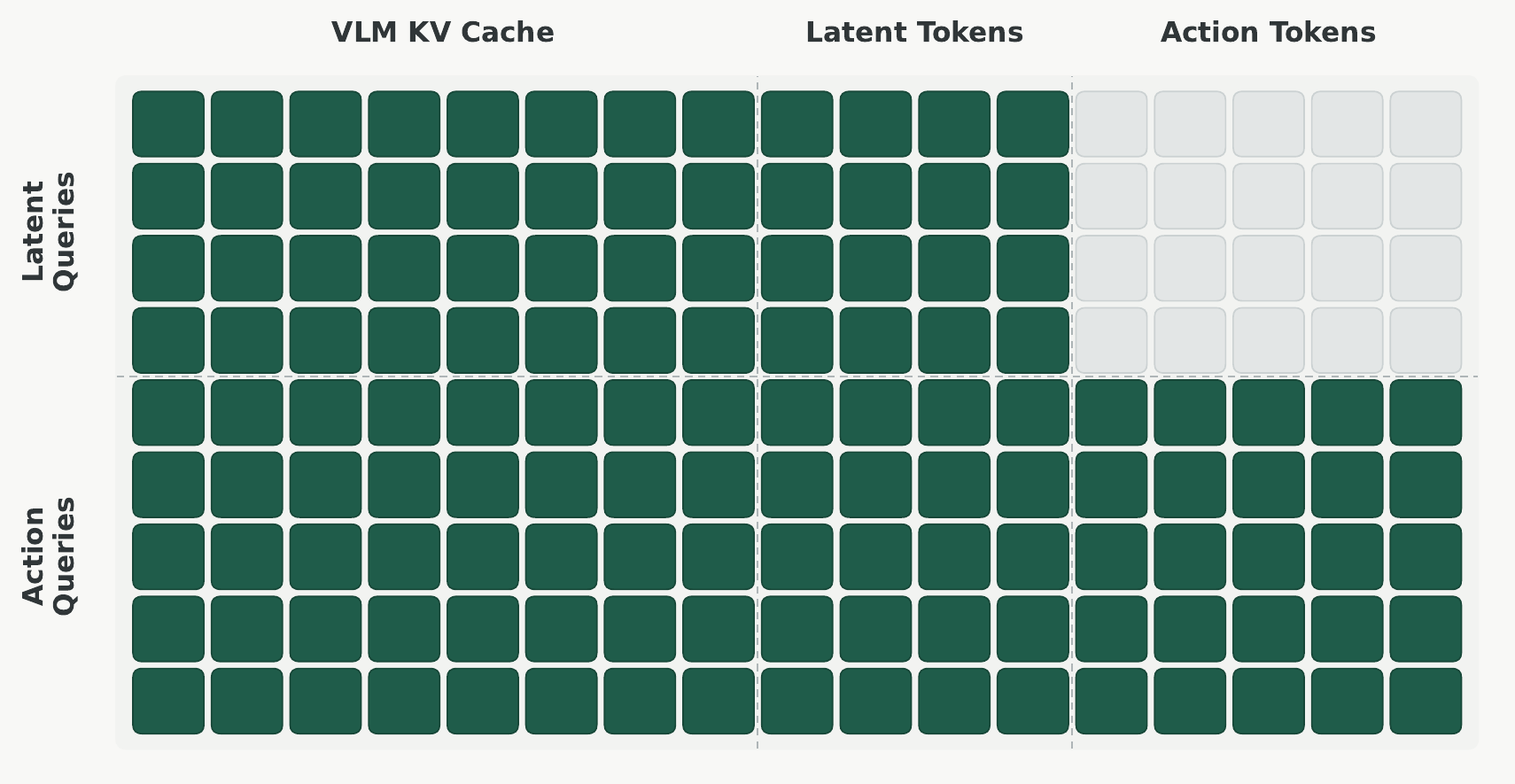}
    \caption{
    Attention mask used in MindWPI and MindLWPI.
    Rows denote query tokens and columns denote key/value tokens.
    Latent queries can attend to the VLM KV cache and latent tokens, but are masked from action tokens to prevent shortcut prediction.
    Action queries can attend to the VLM KV cache, latent tokens, and action tokens, allowing action generation to use predictive latent context.
    }
    \label{fig:attention_mask_en}
\end{figure}
All action-expert tokens can attend to the VLM cache. In MindPI and MindLPI, the action-expert sequence contains only action tokens, and action tokens are mutually visible. In MindWPI and MindLWPI, the action-expert sequence contains latent tokens and action tokens. To prevent future latent prediction from exploiting the action trajectory as a shortcut, latent tokens cannot attend to action tokens; action tokens can attend to both latent tokens and other action tokens. In other words, latent prediction can rely only on current vision-language context and current latent representations, while action generation can use predictive latent context.

DiT blocks inject the flow-matching timestep condition using AdaLN: the timestep is encoded with sinusoidal positional encoding and an MLP into shift, scale, and gate parameters for attention and feed-forward layers. Action tokens use RoPE positions and are appended after the VLM input sequence.

\section{MindLPI Language Supervision Details}
\label{app:language_action}

MindLPI and MindLWPI use two formats for action verbalization.

\textbf{Discrete integer format.} Action values normalized to $[-1,1]$ are uniformly discretized into $1000$ bins and represented as integer sequences. This format does not depend on physical units and is suitable for data with different action spaces or without a unified physical scale.

\textbf{Natural-language format.} We first extract the $7$-dimensional right-arm action and recover the normalized action to physical units according to the global quantiles $(q_{01},q_{99})$:
\begin{equation}
\mathbf{a}^{\mathrm{phys}}
=\frac{1}{2}(\mathbf{a}^{\mathrm{norm}}+\mathbf{1})\odot(q_{99}-q_{01})+q_{01}.
\end{equation}
We then sum over the $16$-step action chunk, convert translation to centimeters, convert rotation to degrees and round it to $5^\circ$, and determine gripper open/close state according to a threshold. The final text has the form ``move forward 12 cm, move up 8 cm, close gripper.'' When downstream actions are joint-space increments or lack physical units, we use the discrete integer format.

The current language supervision is provided only by LAP-style action-description templates. MindLPI optimizes $\mathcal{L}_{\mathrm{act}}+\lambda_{\mathrm{lang}}\mathcal{L}_{\mathrm{lang}}$ during pre-training, where $\lambda_{\mathrm{lang}}=0.1$. MindLWPI additionally adds the future latent loss and uses the same language loss weight.

\section{MindWPI and MindLWPI Latent Prediction Details}
\label{app:wpi_details}

MindWPI uses a frozen V-JEPA~2 model as the latent feature extractor~\citep{bardes2025vjepa2}. Given current and future frames, it extracts $\mathbf{z}_{\mathrm{cur}},\mathbf{z}_{\mathrm{fut}}\in\mathbb{R}^{B\times N_z\times d_z}$. During pre-training, future frames are sampled with a fixed offset, whose default value is $8$ frames. After freezing the feature extractor, we do not add an external linear projection directly, in order to avoid projecting the features into a low-variance representation; latent encoding and decoding are handled by internal modules of the action expert.

In the currently reported full pre-training experiments for MindWPI/MindLWPI, whenever both action loss and future latent loss are enabled, the pre-training stage uniformly uses $\lambda^{\mathrm{pt}}_{\mathrm{lat}}:\lambda^{\mathrm{pt}}_{\mathrm{act}}=1:1$. The notation $\lambda^{\mathrm{ft}}_{\mathrm{lat}}:\lambda^{\mathrm{ft}}_{\mathrm{act}}=0.1:1$ in the main experiments refers only to downstream fine-tuning. Downstream fine-tuning may choose whether to retain the latent loss depending on the experimental setting.

\textbf{AvgPool-k4 compression.} MindLWPI compresses the $256$ V-JEPA~2 latent tokens with AvgPool-k4 by default: every $4$ adjacent tokens along the token sequence are averaged, resulting in $64$ compressed tokens. This operation is applied to both the current latent and the future latent target. We also tested MLP compression, where an MLP first compresses the hidden dimension to $1/k$ and then concatenates every $k$ tokens into one token. However, experiments show that simple average pooling is already sufficiently effective, so AvgPool-k4 is used by default in the main text.

\section{Training and Fine-tuning Hyperparameters}
\label{app:training_details}

\subsection{Pre-training Data and Sampling}

The OXEMix pre-training data mixture consists of DROID, raw OpenX-Embodiment data, OpenX-Augmented embodiment-augmented data, and the RoboCOIN subset, totaling approximately $5{,}000$ hours \citep{khazatsky2024droid,openx2023,wu2025robocoin}. In the text, OXE raw subset denotes the raw OXE subset including DROID but excluding OXE-Augmented and RoboCOIN. All data are converted into the LeRobot format and mapped into the $14$-dimensional action space. To alleviate oversampling caused by large differences in dataset size, the sampling probability considers both dataset scale and trajectory length. MindWPI and MindLWPI use world-model variants of the same data mixture and additionally provide current and future frames. MindLPI and MindLWPI additionally use action-description text as language supervision.

\subsection{Optimization Settings}

All four paradigms use the AdamW optimizer with $\beta=(0.9,0.95)$. The learning rate is $1\times10^{-5}$ for the VLM backbone and $1\times10^{-4}$ for the action expert. The learning-rate schedule is cosine decay with a minimum learning rate of $5\times10^{-7}$ and a warmup of $5{,}000$ steps. Training uses bfloat16 mixed precision, gradient checkpointing, and gradient clipping with a maximum gradient norm of $1.0$. The maximum number of pre-training steps is $2\times10^5$. Distributed training uses data parallelism and ZeRO-2 optimization \citep{rajbhandari2020zero}. During pre-training, neither the VLM nor the action expert is frozen, except in the MindPI (Frozen VLM) control experiment.

The flow-matching timestep is sampled as $u\sim\mathrm{Beta}(1.5,1.0)$ and transformed by $t=(s_0-u)/s_0$, where $s_0=0.999$. To improve timestep coverage, each sample is replicated $4$ times within a batch, and each replica independently samples $(t,\boldsymbol{\epsilon})$. At inference time, we use $4$ Euler integration steps.

\textbf{Computational resources and pre-training batch size.} The main training runs are conducted on NVIDIA A800 GPUs, each with $80$~GB of memory. Pre-training uses $64$ GPUs with a per-GPU batch size of $8$, yielding a global batch size of $8\times64=512$. The reported batch size counts original training samples before the fourfold replication used to sample independent flow-matching timestep-noise pairs.

\subsection{Downstream Full-Parameter Fine-tuning}

Downstream full-parameter fine-tuning is conducted on standard LIBERO and the corresponding SimplerEnv robot data. LIBERO-Plus is used only for zero-shot robustness evaluation: the 100k-step checkpoint fine-tuned on standard LIBERO is evaluated directly under LIBERO-Plus perturbations without any additional training or adaptation. The model architecture remains consistent with pre-training so that checkpoints can be loaded directly. Taking LIBERO as an example, the native Franka single-arm $7$-dimensional joint-space increment is zero-padded into the $14$-dimensional action space, and a validity mask marks the true action slots.

LIBERO fine-tuning uses $8$ GPUs with a per-GPU batch size of $16$, yielding a global batch size of $16\times8=128$. SimplerEnv fine-tuning uses $16$ GPUs with the same per-GPU batch size of $16$, yielding a global batch size of $16\times16=256$. LIBERO-Plus does not introduce a separate training configuration because it directly evaluates the LIBERO-fine-tuned checkpoint without further adaptation. Fine-tuning is full-parameter by default, uses a maximum of $1\times10^5$ steps, and has a warmup of $2{,}000$ steps. Other optimization settings are the same as in pre-training.

\section{Measured Inference Cost}
\label{app:latency}

We profile batch-one inference on a single NVIDIA RTX 5090 using bfloat16 precision and \emph{torch.compile}. After 20 warm-up cases, timings are averaged over 100 LIBERO observations containing front- and wrist-camera inputs. Table~\ref{tab:latency_report} reports end-to-end latency together with the VLM, V-JEPA, and action-expert components. Relative to MindPI, MindWPI increases median latency from 74.54 to 129.82~ms and peak allocated memory from 11,668 to 13,123~MB. AvgPool-k4 reduces the latent-token count but does not yield a wall-clock speedup in this batch-one profile. These measurements characterize the reported LIBERO input workload; SimplerEnv can use a different number of image tokens and should not be assigned the same hardware-independent latency.

\begin{table}[H]
\centering
\caption{Batch-one inference profile on an RTX 5090 (latency in ms, peak memory in MB). Measurements use bfloat16, \emph{torch.compile}, 20 warm-up cases, and 100 LIBERO observations.}
\label{tab:latency_report}
\scriptsize
\setlength{\tabcolsep}{2.5pt}
\begin{adjustbox}{max width=\textwidth}
\begin{tabular}{lrrrrrrr}
\toprule
Method & Mean & Median & P95 & VLM & V-JEPA & Action & Peak MB \\
\midrule
MindPI & 81.07 & 74.54 & 78.15 & 45.26 & -- & 32.25 & 11668 \\
MindWPI & 130.93 & 129.82 & 135.98 & 38.76 & 26.67 & 62.09 & 13123 \\
MindWPI, AvgPool-k4 & 134.00 & 132.24 & 142.78 & 39.48 & 27.09 & 62.31 & 12996 \\
MindLWPI, AvgPool-k4 & 131.51 & 130.59 & 135.86 & 38.73 & 27.06 & 62.59 & 12996 \\
\bottomrule
\end{tabular}
\end{adjustbox}
\end{table}

\section{Low-Rank Adaptation Settings}
\label{app:lora}

Considering the storage and computational cost of full-parameter fine-tuning, we additionally evaluate low-rank adaptation with LoRA \citep{hu2021lora}. We inject low-rank modules only into the query and value projection matrices of attention layers and compare three injection strategies.

\textbf{Action-side LoRA.} We freeze the VLM and inject LoRA only into the query and value projections of attention layers in the action expert.

\textbf{VLM-side LoRA.} The action expert is fully trainable, and LoRA is injected only into the query and value projections of attention layers in the VLM language backbone. This configuration introduces approximately $94.37\mathrm{M}$ trainable parameters when $r=256$.

\textbf{Both-side LoRA.} LoRA is injected into both the VLM and action expert to evaluate adaptation gains under a larger parameter budget.

These settings compare whether the model can retain the transfer advantages of pre-training under different trainable-parameter budgets. The LoRA results are reported as an efficient-adaptation experiment and are not a primary variable in the fair comparison of the four training paradigms. Table~\ref{tab:lora_raw_en} reports complete results for the four LIBERO suites.

\end{document}